\documentclass{article} % For LaTeX2e
\usepackage{iclr2025_conference,times}

\usepackage{amsmath,amsfonts,bm}

\def\eqref#1{equation~\ref{#1}}
\def\1{\bm{1}}

\DeclareMathAlphabet{\mathsfit}{\encodingdefault}{\sfdefault}{m}{sl}
\SetMathAlphabet{\mathsfit}{bold}{\encodingdefault}{\sfdefault}{bx}{n}

\usepackage{booktabs}
\usepackage{hyperref}
\usepackage{fontawesome} 
\usepackage{makecell}
\usepackage{wrapfig}  
\usepackage{adjustbox}
\usepackage{mathrsfs}
\usepackage{listings} 
\usepackage{pifont}
\usepackage{graphicx}
\usepackage{subfigure}
\usepackage{alltt}  
\usepackage{subcaption}
\usepackage{url}
\usepackage{svg}
\usepackage{makecell}
\usepackage{cleveref}
\crefname{section}{§}{§§}
\Crefname{section}{§}{§§}
\usepackage{svg}
\usepackage[normalem]{ulem} 
\usepackage{colortbl}
\usepackage{multirow}
\usepackage{algorithm}
\usepackage{algpseudocode}
\usepackage[figuresright]{rotating}
\usepackage[most]{tcolorbox}
\definecolor{BoxBackground}{RGB}{240, 240, 240} % 浅灰色背景
\definecolor{BoxFrame}{RGB}{0, 0, 0} % 黑色边框
\definecolor{TitleBackground}{RGB}{0, 0, 0} % 标题背景颜色
\definecolor{TitleText}{RGB}{255, 255, 255} % 标题文字颜色
\tcbset{
  academicbox/.style={
    boxsep=5pt,
    left=2pt,
    right=2pt,
    bottom=0.5pt,
    boxrule=0.5pt,
    colback=BoxBackground,
    colframe=BoxFrame,
    colbacktitle=TitleBackground,
    coltitle=TitleText,
    enhanced,
    attach boxed title to top left={yshift=-0.1in,xshift=0.1in},
    boxed title style={boxrule=0pt,colframe=white},
    title={#1},
  }
}
\newtcolorbox{AcademicBox}[1][]{academicbox=#1}

\definecolor{s_doc_qa_c}{HTML}{da0d68}
\definecolor{m_doc_qa_c}{HTML}{da1d23}
\definecolor{summarization_C}{HTML}{ebb40f}
\definecolor{dialogue_c}{HTML}{187a2e}
\definecolor{synthetic_c}{HTML}{0aa3b5}

\title{L-CiteEval: Do Long-Context Models Truly Leverage Context for Responding?}

\author{Zecheng Tang$^{1}$,\quad Keyan Zhou$^{1}$\thanks{Equal Contribution},\quad Juntao Li$^{1}$\thanks{Corresponding Author},\quad Baibei Ji$^{1}$,\quad Jianye Hou$^{2}$,\quad Min Zhang$^{1}$\\
$^{1}$Soochow University \quad $^{2}$CUHK \\
\texttt{\{zctang, kyzhou, bbjidbj\}@stu.suda.edu.cn},\\ \texttt{\{ljt, minzhang\}@suda.edu.cn},\quad \texttt{jianyehou@link.cuhk.edu.cn} \\
}

\iclrfinalcopy % Uncomment for camera-ready version, but NOT for submission.
\begin{document}

\maketitle

\begin{center}
    \textbf{\textit{\faGithub~Code \& Data: \textcolor{violet}{ \url{https://github.com/ZetangForward/L-CITEEVAL.git}}}}
\end{center}

\begin{abstract}
Long-context models~(LCMs) have made remarkable strides in recent years, offering users great convenience for handling tasks that involve long context, such as document summarization.
As the community increasingly prioritizes the faithfulness of generated results, merely ensuring the accuracy of LCM outputs is insufficient, as it is quite challenging for humans to verify the results from the extremely lengthy context.
Yet, although some efforts have been made to assess whether LCMs respond truly based on the context, these works either are limited to specific tasks or heavily rely on external evaluation resources like GPT-4.
In this work, we introduce \textbf{\textit{L-CiteEval}}, a comprehensive multi-task benchmark for long-context understanding with citations, aiming to evaluate both the understanding capability and faithfulness of LCMs.
L-CiteEval covers 11 tasks from diverse domains, spanning context lengths from 8K to 48K, and provides a fully automated evaluation suite.
Through testing with 11 cutting-edge closed-source and open-source LCMs, we find that although these models show minor differences in their generated results, open-source models substantially trail behind their closed-source counterparts in terms of citation accuracy and recall.
This suggests that current open-source LCMs are prone to responding based on their inherent knowledge rather than the given context, posing a significant risk to the user experience in practical applications.
We also evaluate the RAG approach and observe that RAG can significantly improve the faithfulness of LCMs, albeit with a slight decrease in the generation quality.
Furthermore, we discover a correlation between the attention mechanisms of LCMs and the citation generation process.
\end{abstract}

\section{Introduction}
\label{sec:intro}
The rapid development of Long-context Models~(LCMs) provides users with numerous conveniences in resolving long-context real-world tasks, such as code analysis~\citep{zhu2024deepseek} and long document summarization~\citep{reid2024gemini}. 
Recently, the community has gradually intensified its efforts to enhance the faithfulness of generative artificial intelligence~\citep{manna2024faithfulness}, which is of paramount importance for LCMs.
This is because tasks that involve long context usually require LCMs to respond based on the provided context rather than relying solely on models' intrinsic knowledge. 
Therefore, there is an urgent need for a benchmark to verify whether LCMs truly leverage context for responding and reflect those models' capability on long-context tasks.

\begin{figure}[t]
    \centering
    \includegraphics[width=0.95\linewidth]{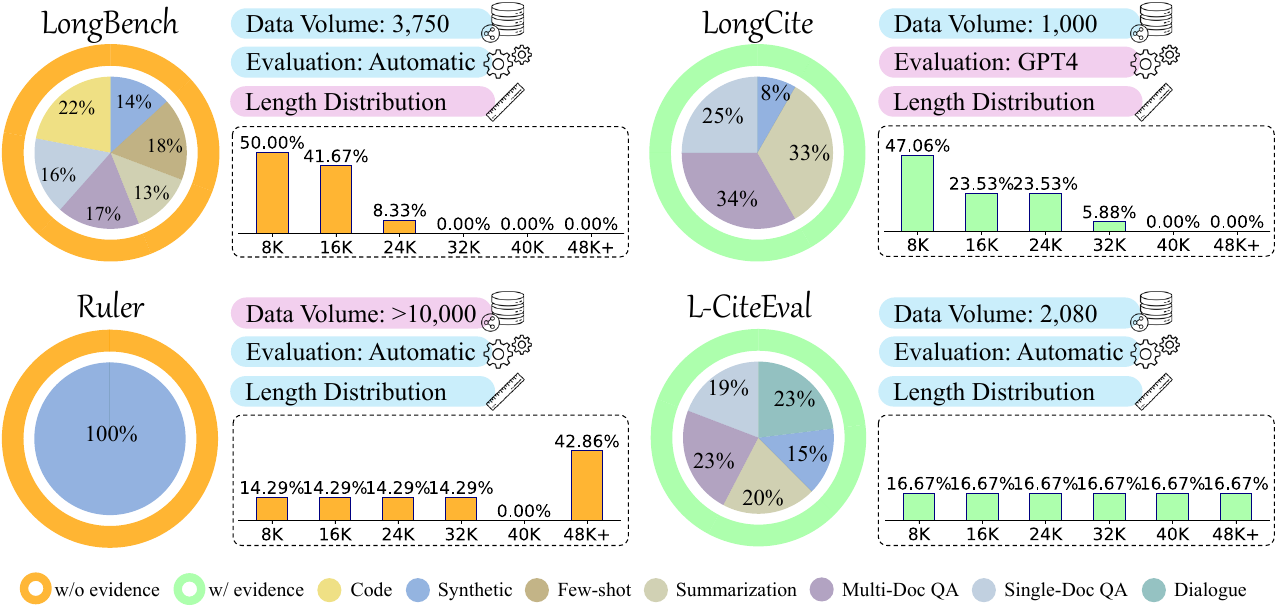}
    \caption{Overview and comparison among different representative benchmarks for LCMs.}
    \vspace{-2em}
    \label{fig:intro}
\end{figure}

To date, substantial efforts have been made to develop benchmarks for evaluating LCMs. 
These endeavors aim to achieve several key objectives:
(1) ensuring that the benchmarks include a \textbf{comprehensive} range of task scenarios and varying context lengths; 
(2) employing automated metrics to guarantee the \textbf{reproducibility} of evaluations; 
(3) incorporating an appropriate volume of test data to maintain evaluation \textbf{efficiency}; and 
(4) offering sufficient \textbf{interpretability}~(e.g., providing evidence to support the responses).
As shown in Fig.~\ref{fig:intro}, taking three representative long-context benchmarks as examples: 
LongBench~\citep{bai2023longbench} primarily evaluates the accuracy of LCMs' responses across a range of realistic and synthetic tasks, with a context length of up to 24K tokens; 
Ruler~\citep{hsieh2024ruler} focuses on using synthetic data to test LCMs' capabilities in information retrieval over long sequences, with context lengths exceeding 48K tokens;
and LongCite~\citep{bai2024longcite} assesses whether models respond based on the content within the context, employing GPT-4 as a judge.
These benchmarks, based on their purpose, can be roughly divided into two categories: (1) evaluating long-context understanding and (2) assessing model faithfulness.
% These benchmarks can be roughly categorized into two groups: testing long-context understanding and evaluating model faithfulness.
The former evaluates model outputs using large volumes of test data to infer LCMs' capabilities but lacks interpretability to the generated results. 
The latter are mainly based on short-context datasets~(e.g., in LongCite, the maximum sequence length only reaches 32K, comprising just 5.88\% of the benchmark) and rely on external resources like GPT-4 to judge faithfulness, making the evaluation results hard to reproduce.
In this work, we introduce \textbf{\textit{L-CiteEval}}, a comprehensive multi-task benchmark for long-context understanding with citations.
As shown in Fig.~\ref{fig:l_citeeval}, given the question and long reference context, L-CiteEval requires LCMs to generate both the statements and their supporting evidence~(citations).
There are \textbf{5} major task categories, \textbf{11} different long-context tasks, with context lengths ranging from \textbf{8K} to \textbf{48K} in L-CiteEval.
To address the timeliness and the risk of data leakage in testing~\citep{ni2024training,apicella2024don}, we incorporate 4 latest long-context tasks as the subsets in L-CiteEval, ensuring that the evaluation remains up-to-date and robust.
Different from previous benchmarks for long-context understanding that primarily assess LCMs based on their predicted answers, L-CiteEval evaluates model performance based on both the generation quality~(whether the predicted answer is correct) and citation quality~(whether the provided citations can support the corresponding answer).
To extend the context length of short-context data, we design a rigorous data construction pipeline to extend the sequence length and mitigate the perturbation introduced from the additional context.
Additionally, to facilitate the ease of use and ensure reproducibility, L-CiteEval offers an automatic evaluation suite.
Considering that the prediction from LCMs can be influenced by both the task difficulty and the context length, we propose two benchmark variants: \textbf{\textit{L-CiteEval-Length}} and \textbf{\textit{L-CiteEval-Hardness}}. 
These two variants strictly control the variables within the evaluation, focusing solely on context length and task difficulty to assess LCMs' capabilities.

% We test 11 cutting-edge and widely-used LCMs, encompassing 3 closed-source LCMs~(e.g., GPT-4o~\citep{gpt4o}) and 8 open-source models featuring different sizes~(e.g., Phi-3.5-mini~\citep{abdin2024phi} and Llama3-ChatQA-2-70B~\citep{xu2024chatqa}) and architectures~(e.g., Qwen2-MoE~\citep{qwen2} and Llama-3.1-Instruct~\citep{llama3modelcard}).
We test 11 cutting-edge and widely-used LCMs, including 3 closed-source and 8 open-source models, which feature different sizes and architectures.
We also explore whether the Retrieval-Augmented Generation~(RAG) technique can improve the faithfulness of LCMs.
Evaluation results indicate that there is a minor difference between open-source and closed-source models regarding generation quality, while open-source models substantially trail behind their closed-source counterparts in terms of citation quality.
Utilizing the RAG technique exhibits a notable improvement in the faithfulness of open-source models, but it slightly impacts the generation quality.
Furthermore, we reveal a correlation between the model's citation generation process and its attention mechanism~(i.e., retrieval head~\citep{wu2024retrieval}), demonstrating the validity of our benchmark and offering insights for future evaluations of LCM faithfulness and the development of advanced LCMs.

\section{Related Works}
\label{sec:related_work}
\subsection{Long-context Understanding Benchmarks}
\label{subsec:lcb}
Currently, there is a growing body of work dedicated to evaluating the long-context understanding capabilities of LCMs.
The majority of benchmarks for LCMs are built based on real-world tasks that inherently encompass long context, including but not limited to long-document QA, long-document summarization, and long-term conversations~\citep{li2023loogle,shaham2023zeroscrolls,an2023eval,LTM,bai2023longbench,dong2023bamboo,zhang2024bench,lee2024can,levy2024same}.
Recently, InfiniteBench~\citep{zhang2024infty} has pushed the boundaries of benchmarks based on real-world tasks by extending the context length beyond 100K tokens.
However, real-world tasks exhibit a variety of forms and evaluation methods, and existing evaluations are applied inconsistently across different works.
Additionally, the generated results can also be influenced by the intrinsic knowledge of LCMs.
To make evaluations more controllable and eliminate the influence of the LCMs' intrinsic knowledge, synthetic benchmarks are often employed~\citep{hsieh2024ruler}.
Among those synthetic benchmarks, task formats can be custom-defined into various types, such as retrieval-based tasks that require the model to extract specific information from a long context~\citep{NIAH,mohtashami2023landmark,xiao2024infllm,liu2024lost,wang2024augmenting}, many-shot in-context learning~\citep{agarwal2024many,bertsch2024context}, fact reasoning~\citep{kuratov2024babilong,karpinska2024one}, \textit{etc}.
In this work, we introduce L-CiteEval, which contains both real-world tasks and synthetic tasks for long-context understanding with citations.
By requiring LCMs to provide evidence to support their predictions, we can also mitigate the challenge of being unable to test whether LCMs respond based on their intrinsic knowledge or the provided context.

\begin{figure}[t]
    \centering
    \vspace{-1em}
    \includegraphics[width=0.9\textwidth]{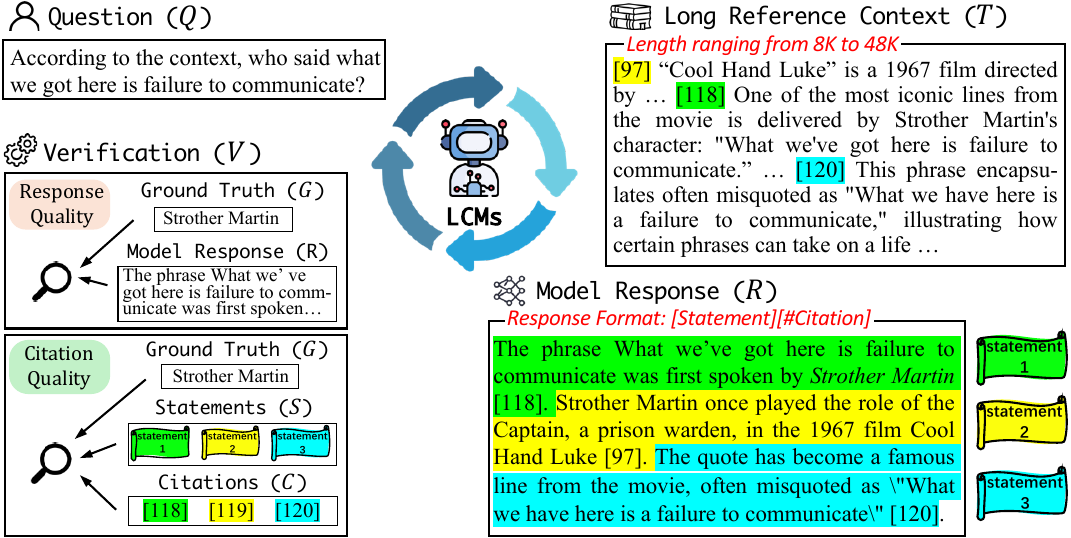}
    \caption{Task format and pipeline of L-CiteEval benchmark.}
    \label{fig:l_citeeval}
    \vspace{-1em}
\end{figure}
\subsection{Citation Generation}
\vspace{-0.5em}
\label{subsec:citegen}
The citation generation task aims to verify whether the model predictions are supported by the referenced source~\citep{li2023survey}.
To evaluate the citations generated by models, \citet{rashkin2023measuring} first proposed \textit{attributed to identified sources}~(AIS) evaluation framework to measure the faithfulness of the model outputs.
Then, some works began to improve the AIS framework in different tasks~(such as single-document QA~\citep{bohnet2022attributed} and fact checking~\citep{honovich2022true}) and domains~(such as science~\citep{funkquist2022citebench} and commerce~\citep{liu2023evaluating}).
To enhance the evaluation precision of citations within the generated text, \citet{qian2023webbrain,kamalloo2023hagrid,li2023towards} made great contributions based on the QA tasks.
With the advancement of generative AI, citation generation has begun to require models themselves to generate citations that support their predictions~\citep{gao2023enabling}.
More recently, \citet{bai2024longcite} introduced \textit{LongCite}, which represents the first attempt at citation generation in long context question-answering tasks.
Compared with LongCite, L-CiteEval is (1) more comprehensive -- it covers a wider range of tasks, supports longer context lengths, and strictly categorizes tasks by length intervals; (2) more reproducible -- it relies entirely on automatic evaluation metrics without reliance on GPT-4 or human judgments; and (3) more efficient -- the task and data distribution are well-designed in L-CiteEval, enabling users to utilize a limited amount of testing data to reflect the LCMs' capabilities.
\section{L-CiteEval: Task and Construction}
\label{sec:bench_construct}

\subsection{Problem Definition and Evaluation Metrics}
\label{subsec:problem_definition}
\paragraph{Problem Definition}
As shown in Fig.~\ref{fig:l_citeeval}, given the long context $T$ and question $Q$, a LCM is expected to generate the response $R$, which contains several statements $\mathcal{S}=\{s_1, s_2, \cdots, s_{n}\}$ and their corresponding citations~$\mathcal{C}=\{c_1, c_2, \cdots, c_n\}$. 
The context $T$ is divided into chunks of varying lengths based on the specific task, with each chunk representing a citation segment. 
Specifically, we set large citation chunk sizes for information-concentrated tasks like Single-Document QA to ensure segment integrity while using small citation chunk sizes for information-dispersed tasks like summarization to maximize the number of citations that LCMs can leverage to support the generated results.
The model can then utilize these citation segments to support the statement $s_i$ within the response. 
In terms of output format, we require each statement $s_i$ to be strictly followed by a supporting citation chunk index $c_i$, which can also serve as an enclosure.

\paragraph{Automatic Evaluation}
During the verification stage, the model response is evaluated from two aspects: the response quality and citation quality.
As shown in Tab.~\ref{tab:data_statistic}, for response quality, we employ different evaluation metrics tailored to each specific task, e.g., Precision~(Prec.) and Recall~(Rec.) for QA tasks and Rouge-L~\citep{lin2004rouge} for summarization tasks.
As for citation quality, following~\citet{gao2023enabling}, we adopt Citation Recall~($CR$) to reflect whether the model statements are fully supported by the citations; Citation Precision~($CP$) to detect irrelevant citations; and citation $F_1$ score to represent the overall citation performance.
Besides, we report citation number $N$ to show how many citations the model uses to support its output.
Different from previous works that utilize an NLI model~\citep{gao2023enabling} to automatically determine whether citations support the corresponding statements, we adopt a long-context NLI model deberta-base-long-nli~\citep{sileo-2024-tasksource}, to better align with long-context scenarios.
We describe the calculation of $CR$ and $CP$ in Appendix~\ref{appdix:cit_pre_recall_cal}.

\begin{table}[t]
    \centering
    \caption{Statistic of tasks in L-CiteEval. The citation chunk size for each task is \textit{\{task\}:\{size\}}.}
    \vspace{-0.5em}
    \resizebox{1\textwidth}{!}{
    \begin{tabular}{ l l c c c c c c c c}
        \toprule
         \multirow{2}{*}{\bf Tasks} & \multirow{2}{*}{\bf Source} & \multirow{2}{*}{\bf \makecell[c]{Evaluation \\ Metric}} &  \multicolumn{6}{c}{\bf Length Distribution} & \multirow{2}{*}{\bf Total} \\
         \cmidrule{4-9}
         & & & \bf 0$\sim$8k & \bf 8$\sim$16k & \bf 16$\sim$24k & \bf 24$\sim$32k & \bf 32$\sim$40k & \bf 40$\sim$48k \\
         \midrule
         \rowcolor{s_doc_qa_c!25} \multicolumn{10}{c}{\textit{\textbf{Single-document QA}}~~~(NarrativeQA: 256, Natural Questions: 256)}  \\
         NarrativeQA & \citep{kovcisky2018narrativeqa} & Prec., Rec. & 40 & 40 & 40 & 40 & 40 & 40 & 240\\
         Natural Questions & \citep{kwiatkowski2019natural} & Prec., Rec. & - & - & 40 & 40 & 40 & 40 & 160\\
         \midrule
         \rowcolor{m_doc_qa_c!25} \multicolumn{10}{c}{\textit{\textbf{Multi-document QA}}~~~(HotpotQA: 128, 2WikiMultihopQA: 128)} \\
         HotpotQA & \citep{yang2018hotpotqa} & Prec., Rec. & 40 & 40 & 40 & 40 & 40 & 40 & 240\\
         2WikiMultihopQA & \citep{ho2020constructing} & Prec., Rec. & 40 & 40 & 40 & 40 & 40 & 40 & 240\\
         \midrule
         \rowcolor{summarization_C!25} \multicolumn{10}{c}{\textit{\textbf{Summarization}}~~~(MultiNews: 128, GovReport: 128, QMSum: 128)} \\
         MultiNews & \citep{ghalandari2020large} & Rouge-L & 20 & 20 & 20 & 20 & 20 & - & 100\\
         GovReport & \citep{huang2021efficient} & Rouge-L & 40 & 40 & 40 & 40 & 40 & 40 & 240\\
         QMSum & \citep{zhong2021qmsum} & Rouge-L & 20 & 20 & 20 & 20 & - & - & 80\\
         \midrule
         \rowcolor{dialogue_c!25} \multicolumn{10}{c}{\textit{\textbf{Dialogue Understanding}}~~~(LoCoMo: 256, DialSim: 256)} \\
         LoCoMo & \citep{maharana2024evaluating} & Prec., Rec. & 40 & 40 & 40 & 40 & 40 & 40 & 240\\
         DialSim & \citep{kim2024dialsim} & Prec., Rec. & 40 & 40 & 40 & 40 & 40 & 40 & 240\\
         \midrule
         \rowcolor{synthetic_c!25} \multicolumn{10}{c}{\textit{\textbf{Synthetic Task}}~~~(NIAH: 256, Counting Stars: 128)} \\
         NIAH & \citep{NIAH} & Rouge-1 & 20 & 20 & 20 & 20 & 20 & 20 & 120\\
         Counting Stars & \citep{song2024counting} & Accuracy & 30 & 30 & 30 & 30 & 30 & 30 & 180\\
         \bottomrule
    \end{tabular}}
    \vspace{-1em}
    \label{tab:data_statistic}
\end{table}

\subsection{Benchmark Construction}
\label{subsec:data_collection}
\begin{figure}[t]
    \centering
    \includegraphics[width=1\linewidth]{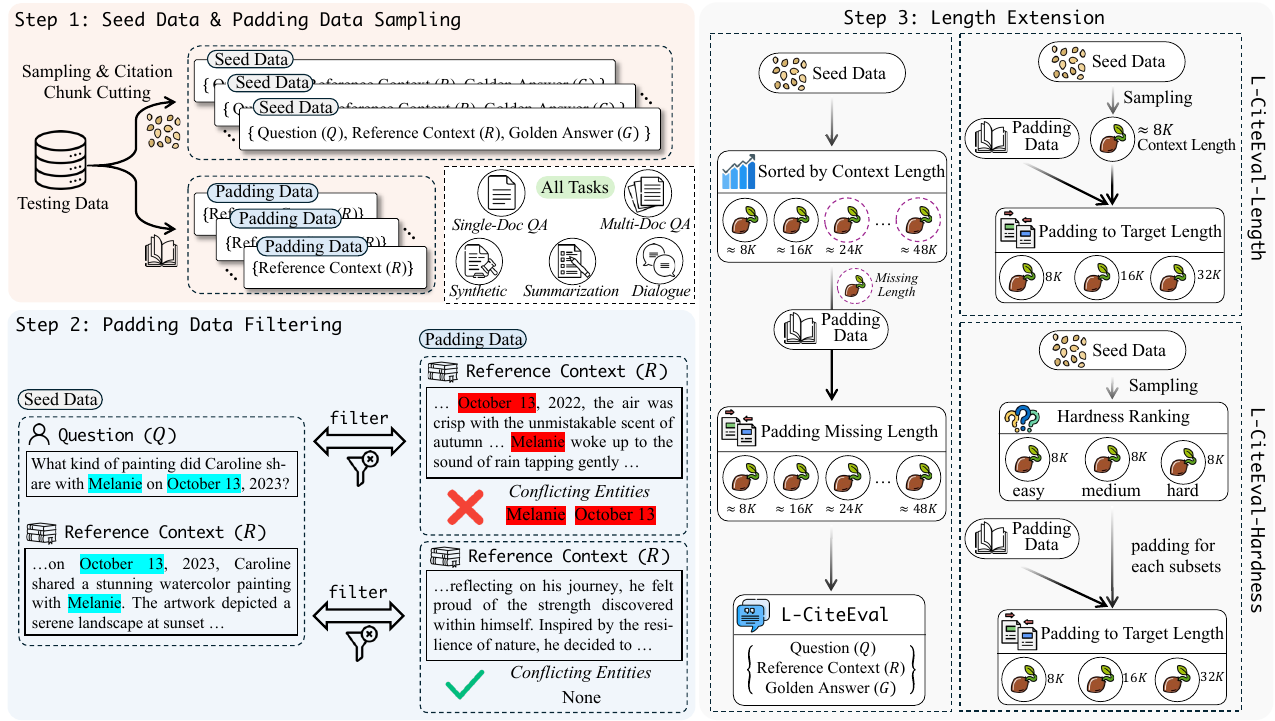}
    \caption{Benchmark construction pipeline.}
    \label{fig:L_CiteEval_construct}
    \vspace{-0.5em}
\end{figure}

There are 5 main categories in the L-CiteEval benchmark: Single-document QA, Multi-document QA, Summarization, Dialogue understanding, and Synthetic tasks, covering both realistic and synthetic tasks.
We report the data source for each task in Table~\ref{tab:data_statistic},
For each task, we utilized the same construction process to handle the dataset.
As shown in Fig.~\ref{fig:L_CiteEval_construct}, the construction process for each task in the L-CiteEval benchmark consists of 3 steps, including (1) Seed Data \& Padding Data Sampling, (2) Padding Data Filtering, and (3) Length Extension.

\paragraph{Step1: Seed Data \& Padding Data Sampling}
Considering the large amount of data in each source dataset, we first sample a portion of testing dataset $\mathcal{D}_{seed}$ as the seed data, from which we can subsequently construct the benchmark.
However, some source datasets, e.g., LoCoMo~\citep{maharana2024evaluating}, exhibit short context.
Consequently, we sample data from the remaining source dataset to serve as the candidate padding data $\mathcal{D}_{pad}$ for length extension.
We divide all the sampled data~($\mathcal{D}_{seed}$ and $\mathcal{D}_{pad}$) into citation chunks of approximately equal size, with sentences as the basic unit.
As mentioned above, we utilize different citation chunk sizes for different tasks.
For tasks involving concentrated information, e.g., single-document QA, we employ smaller chunk sizes, while for tasks involving dispersed information, e.g., summarization, we use larger chunk sizes. 
We report the citation chunk size for each dataset in Table~\ref{tab:data_statistic}.

\paragraph{Step2: Padding Data Filtering}
Using padding data to extend the length of the short-context dataset would introduce additional contextual information and could potentially influence the generated results.
Therefore, we eliminate the padding data that might affect the predictions based on overlapping entities in the text.
Specifically, we apply spaCy\footnote{\url{https://spacy.io/usage/models}}, a Named Entity Recognition model $f_{\theta}$, to extract all the entities $E$ from the question~($E^{(Q)}_{seed}$) and reference context~($E^{(T)}_{seed}$) in $\mathcal{D}_{seed}$, as well as the entities from the reference context~($E^{(T)}_{pad}$) in padding data.
Then, we keep the padding samples $\mathcal{D}_{pad}^{*}$ that share a small entity overlaps with the seed data, which can be written as:
\begin{equation}
\mathcal{D}_{pad}^{*} = \left\{\mathcal{D}_{pad}^{\prime}\mid \mathcal{D}_{pad}^{\prime}\sim\mathcal{D}_{pad},~~E_{pad}^{(T)}=f_{\theta}(\mathcal{D}_{pad}^{\prime}),~~|E_{seed}^{(T)}\cap E_{seed}^{(Q)}\cap E_{pad}^{(T)}|\leq \delta \right\},
\label{equ:pad_data_filter}
\end{equation}
where $\delta$ is the threshold to control the entity overlap between seed data and padding data.
In this paper, we set this $\delta=5$ as a strict criterion to filter out data that may potentially impact the results.

\paragraph{Step3: Length Extension}
After obtaining the padding data $\mathcal{D}_{pad}^{*}$, we leverage these data to extend the context length of seed data $\mathcal{D}_{seed}$.
As shown in Figure~\ref{fig:L_CiteEval_construct}, we have three different benchmark settings, including L-CiteEval and its two variants: L-CiteEval-Quality and L-CiteEval-Length.
Specifically, for the L-CiteEval benchmark setting, given the target length interval of the dataset, we first sort the data according to the context length within each task. 
We then randomly sample contexts from $\mathcal{D}_{pad}^{*}$ to extend the context length and fill in the missing target length intervals.
% Specifically, for the L-CiteEval benchmark setting, given the target length interval of the dataset, we first sort the data according to the context length within each task, and randomly sample context from $\mathcal{D}_{pad}^{*}$ to extend the context length to supplement the missing target length interval.
The L-CiteEval benchmark is designed to benchmark the models comprehensively. Thereby, the seed data and context extension data for all samples are different.
For the L-CiteEval-Length benchmark, which aims to test the model's performance from the context length perspective, we use the same set of seed data and different sets of padding data to extend to various context lengths.
For the L-CiteEval-Hardness benchmark that is designed to benchmark models based on question difficulty, we first quantify and rank the difficulty of each question according to the model's generation quality\footnote{Specifically, we categorize the difficulty level of each sample based on GPT-4o because GPT-4o has been proven to exhibit the highest preference similarity with human annotators~\citep{yadav2024towards}.}. 
Then, we categorize the difficulty into three levels: easy (where the model mostly provides correct answers), medium, and hard (where the model mostly produces incorrect answers). 
We use the same padding data to extend the context length for each difficulty level.
We use GPT-4 as the evaluator to classify the sample difficulty, as it shows the best generation quality.

\paragraph{Benchmarks Overview}
For clarity, we list the differences among the three benchmarks below:
\vspace{-0.7em}
\begin{itemize} \itemsep -0.1em
\item \textit{\textbf{L-CiteEval}} is designed to evaluate the comprehensive capabilities~(\uline{generation quality and citation quality}) of LCMs, which is constructed with different seed data~(\uline{varying question difficulty}) and padding data sources~(\uline{varying context}). This benchmark includes 2,080 testing samples, with 11 tasks across 5 categories.
\item \textit{\textbf{L-CiteEval-Length}} is designed to evaluate the LCMs from the context length perspective, which is constructed with the same seed data source~(\uline{same question difficulty}) but different padding data sources~(\uline{varying context}). 
This benchmark consists of 4 tasks across 4 categories, i.e., NarrativeQA~(Single-Doc QA), HotpotQA~(Multi-Doc QA), GovReport~(Summarization), and Counting Stars~(Synthetic task), with each task containing 200 testing samples.
For each task, we establish three context length intervals: 8K, 16K, and 32K.
\item \textit{\textbf{L-CiteEval-Hardness}} is designed to evaluate the LCMs from the task difficulty perspective, which is constructed with the different seed data source~(\uline{varying question difficulty}) but same padding data sources~(\uline{same context}). This benchmark shares the same data distribution and volume with \textit{\textbf{L-CiteEval-Length}}, except that the scoring is based on task difficulty~(Easy, Medium, and Hard) rather than context length.
\end{itemize}

% \paragraph{Benchmark Overview}
% For clarity, we list the differences among the three benchmarks below.
% \begin{itemize} \itemsep -0.1em
% \item \textit{\textbf{L-CiteEval}} is designed to benchmark the comprehensive capabilities~(generation quality and citation quality) of LCMs, which is constructed with different seed data~(\colorbox{green!20}{varying question difficulty}) and padding data sources~(\colorbox{blue!20}{varying context}).
% \item \textit{\textbf{L-CiteEval-Length}} is designed to benchmark the LCMs from the context length perspective, which is constructed with the same seed data source~(\colorbox{green!20}{same question difficulty}) but different padding data sources~(\colorbox{blue!20}{varying context}).
% \item \textit{\textbf{L-CiteEval-Hardness}} is designed to benchmark the LCMs from the question difficulty perspective, which is constructed with the different seed data source~(\colorbox{green!20}{varying question difficulty}) but the same padding data sources~(\colorbox{blue!20}{same context}).
% \end{itemize}

\section{Experiments}
\label{sec:experiments}
\begin{table}[t]
\centering
\small
\caption{Statistic of LCMs. $*$ means the model utilizing YaRN~\citep{peng2023yarn} to extend the base context length. $\dagger$ denotes the MoE model, where activated parameters are enclosed in parentheses.}
\vspace{-0.5em}
\resizebox{\linewidth}{!}{
\begin{tabular}{l c c c c}
\toprule
\bf Model & \bf Ctx. Size & \bf \#Param & \bf Architecture & \bf Open-source \\
\midrule
GPT-4o~(20240513)~\citep{gpt4o} & 128K & \faLock & \faLock & \ding{55} \\ %~(Version: 20240513)
o1-mini~\citep{o1} & 128K & \faLock & \faLock & \ding{55} \\
Claude-3.5-Sonnet~(20240620)~\citep{claude_3_5} & 200K & \faLock & \faLock & \ding{55} \\ %~(Version: 20240620)
\midrule
Qwen2.5-3B-Instruct~\citep{qwen2.5} & 32K (128K$^{*}$) & 3B & Decoder-Only & \ding{51} \\
Phi-3.5-mini-instruct~\citep{abdin2024phi} & 128K & 3.8B & Decoder-Only & \ding{51} \\
Llama-3.1-8B-Instruct~\citep{llama3} & 128K & 8B & Decoder-Only & \ding{51} \\
GLM-4-9B-Chat~\citep{glm2024chatglm} & 128K & 9B & Decoder-Only & \ding{51} \\
Mistral-NeMo-Instruct-2407~\citep{nemo} & 128K & 12B & Decoder-Only & \ding{51} \\
Qwen2-57B-A14B-Instruct~\citep{qwen2} & 32K (128K$^{*}$) & 57B (14B$^{\dagger}$) & MoE & \ding{51} \\
Llama-3.1-70B-Instruct~\citep{llama3} & 128K & 70B & Decoder-Only & \ding{51} \\
Llama3-ChatQA-2-70B~\citep{xu2024chatqa} & 128K & 70B & Decoder-Only & \ding{51} \\
\bottomrule
\end{tabular}}
\label{tab:statistic_lcms}
\vspace{-1em}
\end{table}
We conduct experiments with 11 latest LCMs, including 3 closed-source LCMs and 8 open-source LCMs, each with a context window size of at least 128K tokens, encompassing different parameters~(ranging from 3B to 70B) and model architectures~(dense model and MoE model).
The statistic of LCMs is shown in Tab.~\ref{tab:statistic_lcms}.
We provide one demonstration within the prompt for each task to make the model's output format more standard, i.e., one-shot learning during the inference time, and employ the same instruction for every LCM.
Demonstration of model prediction, question, and instruction for each task is shown in Appendix~\ref{appdix:case_study}.
We benchmark all the LCMs with \textbf{\textit{L-CiteEval}} and then select 6 representative LCMs~(including 1 closed-source LCMs and 4 open-source LCMs) to further evaluate on \textbf{\textit{L-CiteEval-Length}} and \textbf{\textit{L-CiteEval-Hardness}} benchmarks.

\subsection{Model Performance on L-CiteEval}
\label{subsec:model_performance}
We report the citation quality in Tab.~\ref{tab:l_eval_cite_1} ( information-concentrated tasks that require models to seek local information in several citation segments) and Tab.~\ref{tab:l_eval_cite_2} (information-dispersed tasks that require models to seek global information from the entire context) and report the generation quality in Tab.~\ref{tab:main_results_correct}.

\begin{table}[t]
    \centering
    \large
    \caption{Citation quality of LCMs in information-concentrated tasks within L-CiteEval.}
    \vspace{-0.5em}
    \resizebox{\textwidth}{!}{
    \begin{tabular}{l | cccc | cccc | cccc}
        \toprule
         \multirow{2}{*}{\bf Models} &  \multicolumn{4}{c|}{\bf Single-Doc QA } &  \multicolumn{4}{c|}{\bf Dialogue Understanding} &  \multicolumn{4}{c}{\bf Needle in a Haystack } \\
         \cmidrule{2-13}
         & \makecell[c]{$\mathbf{CP}$} & \makecell[c]{\bf $\mathbf{CR}$} & \makecell[c]{\bf $\mathbf{F_1}$} & \makecell[c]{\bf $\mathbf{N}$} & \makecell[c]{$\mathbf{CP}$} & \makecell[c]{\bf $\mathbf{CR}$} & \makecell[c]{\bf $\mathbf{F_1}$} & \makecell[c]{\bf $\mathbf{N}$} & \makecell[c]{$\mathbf{CP}$} & \makecell[c]{\bf $\mathbf{CR}$} & \makecell[c]{\bf $\mathbf{F_1}$} & \makecell[c]{\bf $\mathbf{N}$} \\
         \midrule
          \rowcolor{gray!20} \multicolumn{13}{c}{\textit{\textbf{\faLock~~~Closed-source LCMs}}}  \\
         GPT-4o & 32.05 & \bf 38.12 & 33.48 & 2.02 & 53.90 & \bf 64.25 & \bf 56.76 & 2.17 & \bf 76.25 & \bf 76.67 & \bf 76.39 & 1.12  \\
         Claude-3.5-sonnet & \bf 38.70 & 37.79 & \bf 37.43 & 3.54 & \bf 54.45 & 50.48 & 51.45 & 2.83 & 65.00 & 68.33 & 65.97 & 1.04  \\
         o1-mini & 29.83 & 35.33 & 31.66 & 3.38 &45.54 & 50.74 & 47.21 & 2.63 & 25.42 & 28.33 & 26.25 & 1.58 \\
          \rowcolor{blue!5} \multicolumn{13}{c}{\textit{\textbf{\faUnlock~~~Open-source LCMs}}}  \\
         Qwen2.5-3b-Ins & 7.13 & 5.83 & 6.00 & 1.75 & 9.53 & 9.71 & 8.41 & 2.33 & 12.08 & 12.50 & 12.22 & 1.12   \\
         Phi-3.5-mini-Ins & 21.06 & 20.46 & 19.14 & 2.86 & 20.39 & 24.27 & 20.57 & 2.27 & 11.11 & 12.50 & 11.53 & 1.20 \\
         Llama-3.1-8B-Ins & 22.68 & 24.73 & 22.64 & 2.59 & \underline{51.86} & \bf 57.58 & \underline{53.50} & 2.08 & 34.31 & 35.83 & 34.72 & 0.99 \\
         Glm-4-9B-chat & \bf 29.00 & \bf 28.66 & \bf 28.05 & 2.21 & \bf 54.54 & 55.62 & \bf 53.58 & 1.78 & \underline{46.53} & \bf 50.83 & \bf 47.78 & 1.23  \\
         Mistral-Nemo-Ins & 4.34 & 3.68 & 3.76 & 0.68 & 23.91 & 24.33 & 23.50 & 1.35 & 11.11 & 12.50 & 11.53 & 1.18 \\
         Qwen2-57B-A14B-Ins & 4.90 & 3.43 & 3.82 & 1.27 & 22.63 & 22.54 & 21.61 & 1.80 & 15.28 & 15.83 & 15.42 & 1.17 \\
         Llama-3.1-70B-Ins & \underline{25.89} & \underline{26.89} & \underline{26.11} & 1.23 & 51.71 & \underline{56.20} & 53.19 & 1.76 & \bf 46.67 &  \underline{46.67} & \underline{46.67} & 0.82 \\
         ChatQA-2-70B & 21.75 & 22.54 & 21.92 & 1.12 &  47.67 & 51.25 & 48.77 & 1.29 & 38.33 & 38.33 & 38.33 & 0.95\\
         \bottomrule
    \end{tabular}}
    \vspace{-0.5em}
    \label{tab:l_eval_cite_1}
\end{table}

\begin{table*}[t]
    \centering
    \large
    \caption{Citation quality of LCMs in information-dispersed tasks within L-CiteEval.}
    \vspace{-0.5em}
    \resizebox{\textwidth}{!}{
    \begin{tabular}{l | cccc | cccc | cccc}
        \toprule
         \multirow{2}{*}{\bf Models} &  \multicolumn{4}{c|}{\bf Multi-Doc QA } &  \multicolumn{4}{c|}{\bf Summarization} &  \multicolumn{4}{c}{\bf Counting Stars } \\
         \cmidrule{2-13}
         & \makecell[c]{$\mathbf{CP}$} & \makecell[c]{\bf $\mathbf{CR}$} & \makecell[c]{\bf $\mathbf{F_1}$} & \makecell[c]{\bf $\mathbf{N}$} & \makecell[c]{$\mathbf{CP}$} & \makecell[c]{\bf $\mathbf{CR}$} & \makecell[c]{\bf $\mathbf{F_1}$} & \makecell[c]{\bf $\mathbf{N}$} & \makecell[c]{$\mathbf{CP}$} & \makecell[c]{\bf $\mathbf{CR}$} & \makecell[c]{\bf $\mathbf{F_1}$} & \makecell[c]{\bf $\mathbf{N}$} \\
         \midrule
         \rowcolor{gray!20} \multicolumn{13}{c}{\textit{\textbf{\faLock~~~Closed-source LCMs}}}  \\
         GPT-4o & 57.48 & \bf 58.50 & 56.10 & 1.71 & 34.37 & 54.28 & 41.60 & 22.86 & \bf 83.37 & \bf 81.18 & \bf 81.71 & 4.54 \\
         Claude-3.5-sonnet & \bf 66.85 & 55.62 & \bf 58.58 & 2.44 & \bf 36.70 & \bf 55.03 & \bf 43.45 & 17.70 & 73.01 & 75.83 & 73.15 & 4.81 \\
         o1-mini & 49.95 & 49.60 & 48.58 & 1.78 & 20.23 & 33.61 & 24.83 & 19.58 & 34.06 & 46.46 & 38.45 & 6.73  \\
         \rowcolor{blue!5} \multicolumn{13}{c}{\textit{\textbf{\faUnlock~~~Open-source LCMs}}}  \\
         Qwen2.5-3b-Ins & 13.17 & 8.04 & 9.37 & 1.96 & 7.72 & 12.15 & 9.09 & 9.52 & 3.82 & 1.48 & 2.01 & 1.66 \\
         Phi-3.5-mini-Ins & 11.89 & 10.25 & 10.53 & 1.71 & 10.90 & 10.94 & 9.60 & 8.23 & 4.19 & 3.67 & 4.09 & 3.48 \\
         Llama-3.1-8B-Ins & 43.41 & 42.15 & 41.64 & 1.62 & 19.57 & 23.03 & 20.83 & 18.31  & 16.87 & \underline{18.26} & \underline{19.18} & 4.19  \\
         Glm-4-9B-chat & \underline{47.91} & 44.75 & 45.09 & 1.64 & \bf 29.16 & \bf 37.29 & \bf 31.92 & 11.38 & \underline{18.15} & 15.69 & 16.21 & 4.52 \\
         Mistral-Nemo-Ins & 17.61 & 15.45 & 15.85 & 0.70 & 11.21 & 14.85 & 12.40 & 5.45 & 3.09 & 2.92 & 3.26 & 2.32 \\
        Qwen2-57B-A14B-Ins & 17.30 & 12.07 & 13.61 & 1.06 & 4.01 & 3.37 & 3.19 & 3.81 & 4.37 & 4.37 & 4.24 & 4.24 \\
         Llama-3.1-70B-Ins & \bf 49.64 & \bf 54.02 & \bf 50.74 & 1.42 & \underline{25.50} & \underline{31.99} & \underline{27.91} & 11.78 & \bf 66.85 & \bf 61.74 & \bf 63.73 & 4.37\\
         ChatQA-2-70B & 47.20 & \underline{49.51} & \underline{47.92} & 1.10 & 19.57 & 23.60 & 20.89 & 11.81 & 14.02 & 11.22 & 13.22 & 3.49\\
         \bottomrule
    \end{tabular}}
    \vspace{-1em}
    \label{tab:l_eval_cite_2}
\end{table*}

\subsubsection{Analysis of Citation Quality}
\paragraph{Open-source LCMs versus Closed-source LCMs} 
Overall, there is still a significant performance gap between open-source LCMs and closed-source LCMs~(excluding o1-mini), especially in tasks involving the reasoning step.
Specifically, we can observe that: (1) closed-source LCMs generally provide more accurate citations~(larger $F_1$ score) and tend to cite more segments with the context~(larger value of $N$); (2) in the Dialogue Understanding task, the performance of the strongest open-source LCMs~(Llama-3.1-70B-Instruct) has approached that of the closed-source LCMs. However, in other tasks requiring reasoning, particularly in synthetic tasks, although strong open-source LCMs like GLM-4-9B-Instruct cite a similar number of segments as the closed-source models, the quality of these citations is lower, resulting in a performance gap of nearly 20 $F_1$ points.

\paragraph{Performance of Open-source LCMs}
In general, there is significant room for open-source LCMs to improve, and medium-sized open-source LCMs~(Llama-3.1-8B-instruct and GLM-4-9B-Chat) are highly competitive, with performance that matches or even exceeds that of large LCMs~(Llama-3.1-70B-instruct). 
More concretely, our findings are:
(1) The improvement in citation quality does not directly correlate with the increase in model parameters. As the number of model parameters increases, citation performance does not consistently improve, but overall, large LCMs (70B) perform well, and medium-sized LCMs (8B and 9B) show very promising results; 
(2) The actual activated parameters of LCMs are crucial, as evidenced by the MoE LCM~(Qwen2-57B-A14B) exhibiting significantly lower citation quality, even under-performing small dense LCMs such as Phi-3.5-mini-instruct; 
(3) Training data diversity is essential for LCMs. Taking ChatQA-2-70B, which is primarily trained on QA task datasets, as an example, we can observe that ChatQA-2-70B performs exceptionally well on Single-Doc QA tasks and Multi-Doc QA tasks but struggles significantly with the synthetic tasks and summarization tasks.

\paragraph{Performance of Closed-source LCMs}
Among closed-source LCMs, GPT-4o and Claude-3.5-sonnet demonstrate strong performance on L-CiteEval, with GPT-4o surpassing all the experimental open-source LCMs across all tasks in citation quality. 
Notably, while o1-mini achieves unparalleled results in reasoning tasks such as GSM8K~\citep{cobbe2021training} and Livecodebench~\citep{jain2024livecodebench}, its citation generation capability significantly deteriorates in long-text scenarios. 
Particularly in synthetic tasks and summarization tasks, which require LCMs to search for dispersed key information and use the retrieval information to respond, o1-mini's performance is significantly inferior to strong open-source LCMs, such as Llama-3.1-70B-instruct.
This suggests that the o1-mini model falls short in retrieving key information from the context for responding.

\begin{table}[t]
    \centering
    \small
    \caption{Generation quality of LCMs on L-CiteEval, where $\dagger$ denotes the NIAH results, $\ddagger$ denotes the Counting Stars results, and Summ. denotes the summarization task.}
    \vspace{-0.5em}
    \resizebox{\textwidth}{!}{
    \begin{tabular}{l | cc | cc | c | cc | cc}
        \toprule
         \multirow{2}{*}{\bf Models} & \multicolumn{2}{c|}{\bf Single-Doc QA } &  \multicolumn{2}{c|}{\bf Multi-Doc QA} &  \multicolumn{1}{c|}{\bf Summ.} &  \multicolumn{2}{c|}{\bf Dialogue} &  \multicolumn{2}{c}{\bf Synthetic } \\
         \cmidrule{2-10}
         & \bf Prec.& \bf Rec. & \bf Prec.& \bf Rec. & \bf{Rouge-L} & \bf Prec.& \bf Rec. & \bf{Rouge-1}$^{\dagger}$ & \bf{Acc}$^{\ddagger}$ \\
         \midrule
         \rowcolor{gray!20} \multicolumn{10}{c}{\textit{\textbf{\faLock~~~Closed-source LCMs}}}  \\
         GPT-4o & \bf 11.78 & 70.37 & \bf 10.34 & \bf 87.38 & 20.15 & \bf 9.81 & \bf 65.35 & \bf 89.24 & \bf 91.88 \\
         Claude-3.5-sonnet & 5.96 & \bf 71.96 & 4.30 & 80.77 & \bf 22.06 & 3.71 & 57.80 & 88.33 & 69.65\\
         o1-mini & 10.30 & 66.44 & 7.36 & 64.25 & 19.22 & 7.02 & 54.27 & 54.98 & 57.29\\
         \rowcolor{blue!5} \multicolumn{10}{c}{\textit{\textbf{\faUnlock~~~Open-source LCMs}}}  \\
         Qwen2.5-3b-Ins & 8.91 & 60.28 & 3.82 & 52.41 & \underline{22.39}  & 4.58 & 40.77 & 84.49 & 26.81\\
         Phi-3.5-mini-Ins & 8.62 & 62.34 & 7.82 & 64.54  & 19.48  & 11.39 & 52.77 & 73.83 & 61.32\\
         Llama-3.1-8B-Ins & 10.11 & \bf 68.13 & 7.66 & 68.84  & 20.90  & 11.07 & \underline{58.84} & 85.11 & 33.75\\
         Glm-4-9B-chat & 11.22 & \underline{67.25} & 7.88 & \bf 77.97 & 21.42  & 7.69 & 51.25 & \underline{90.81} & 58.82\\
         Mistral-Nemo-Ins & 10.53 & 59.71 & 8.78 & 67.70 & 20.83  & 9.27 & 49.26 & 87.88 & 18.06\\
         Qwen2-57B-A14B-Ins & 12.93 & 61.71 & \underline{15.25} & 57.53 & \bf 22.95  & 14.32 & 52.23 & \bf 91.30 & 63.61\\
         Llama-3.1-70B-Ins & \underline{15.23} & 67.08 & 12.50 & \underline{76.40} & 22.29  & \underline{19.62} & \bf 62.91 & 88.18 & \bf 89.03\\
         ChatQA-2-70B  & \bf 43.25 & 61.20 & \bf 34.95 & 55.64  & 22.06  & \bf 26.57 & 58.34 & 70.14 &  \underline{78.68}\\
         \bottomrule
    \end{tabular}}
    \label{tab:main_results_correct}
\end{table}
\subsubsection{Analysis of Generation Quality}
From Table~\ref{tab:main_results_correct}, we can find:
(1) In Single-Doc QA, Multi-Doc QA, and Dialogue understanding tasks, closed-source LCMs significantly outperform open-source LCMs in recall scores. 
This indicates that the statements of closed-source LCMs contain the correct answers. 
However, closed-source LCMs tend to generate excessive statements to substantiate the results, consequently leading to lower precision scores.
In Summarization and Synthetic tasks, the gap between closed-source and strong open-source LCMs is small, as the corresponding evaluation results are close, e.g., 22.06 Rouge-L score of Claude-3.5-sonnet versus 22.95 Rouge-L score of Qwen2-57B-A14B-Instruct in Summarization tasks;
(2) Open-source LCMs tend to achieve better performance as the model parameters increase.
Combined with the mediocre citation quality of large LCMs mentioned above, we speculate that larger LCMs rely more on their internal knowledge~(which might include task-specific information) rather than responding based on the provided context. 
Consequently, their outputs are more often drawn from inherent knowledge rather than the context itself.
This finding is also consistent with the current research~\citep{small_hallu}.

\begin{figure}[t]
    \centering
    \subfigure[Model Performance on L-CiteEval-Length.]{
        \includegraphics[width=1\linewidth]{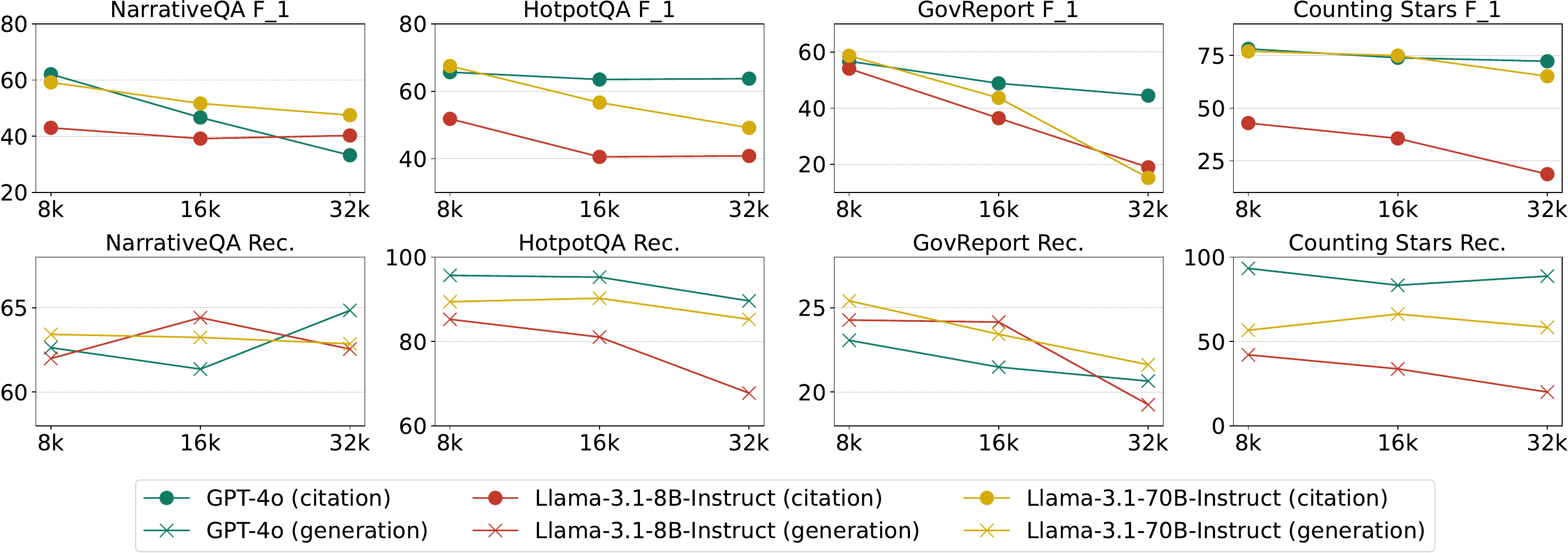}
        \label{fig:result_l_citeeval_length}
    }
    \subfigure[Model Performance on L-CiteEval-Hardness.]{
        \includegraphics[width=1\linewidth]{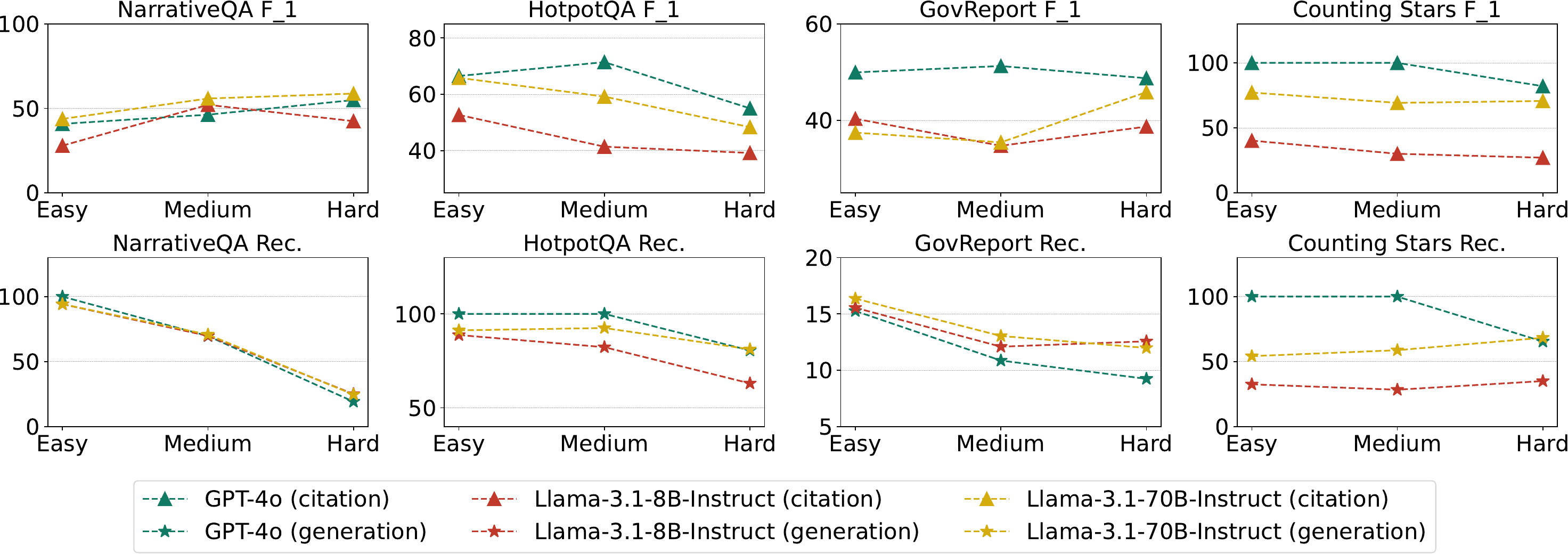}
        \label{fig:result_l_citeeval_hardness}
    }
    \vspace{-0.5em}
    \caption{Model Performance on L-CiteEval-Length and L-CiteEval-Hardness, where we report F\_1 score for citation quality and recall score~(Rec.) for generation quality.}
\end{figure}

\subsection{Model Performance on L-CiteEval-Length and L-CiteEval-Hardness}
\subsubsection{Impact of Context Length for LCMs}
We report the LCMs' performance on L-CiteEval-Length in Fig.~\ref{fig:result_l_citeeval_length}.
When keeping task difficulty constant but extending the context length, we can observe an overall decline in open-source LCMs' performance. 
Specifically, the smallest model, Llama-3.1-8B-Instruct, is the most affected by longer contexts. 
For instance, in the HotpotQA task, its performance drops by around 20 points as the context length increases from 8K to 32K.
Larger models, such as Llama-3.1-70B-Instruct, are slightly impacted.
However, the closed-source LCM~(GPT-4o) maintains a relatively stable performance, showing minimal degradation. 
This suggests that open-source LCMs are more susceptible to irrelevant context, leading to a drop in both generation and faithfulness. 
More details and model performance on L-CiteEval-Length benchmark are shown in Appendix~\ref{appdix:l_citeeval_length}.

\subsubsection{Impact of Task Difficulty for LCMs}
We divide each task into different difficulty levels based on the generation quality of GPT-4o.
The LCMs' performance on L-CiteEval-Hardness is shown in Fig.~\ref{fig:result_l_citeeval_hardness}. 
We observe that as task difficulty increases, the generation quality of LCMs generally decreases~(except for the synthetic task Counting star, which open-source LCMs consistently perform poorly on). 
However, citation quality does not display a consistent trend, though all LCMs demonstrate similar patterns across tasks.
This aligns with our intuition that faithfulness is not strongly correlated with task difficulty.
Besides, these results also underscore a gap between citation quality, which reflects the model's ability to retrieve information from the context, and the generation quality of LCMs.
More details and model performance on L-CiteEval-Hardness benchmark are shown in Appendix~\ref{appdix:l_citeeval_hardness}.
\section{Analysis}
\label{sec:analysis}
Given outstanding performance retrieval-augmented generation~(RAG) on long-context understanding tasks~\citep{li2024retrieval,yu2024defense}, we explore whether RAG can enhance long-context understanding in citation generation tasks.
Furthermore, we will analyze the relevance between the citations produced by LCM and its internal attention mechanisms.

\begin{figure}[t]
    \centering
    \includegraphics[width=1\linewidth]{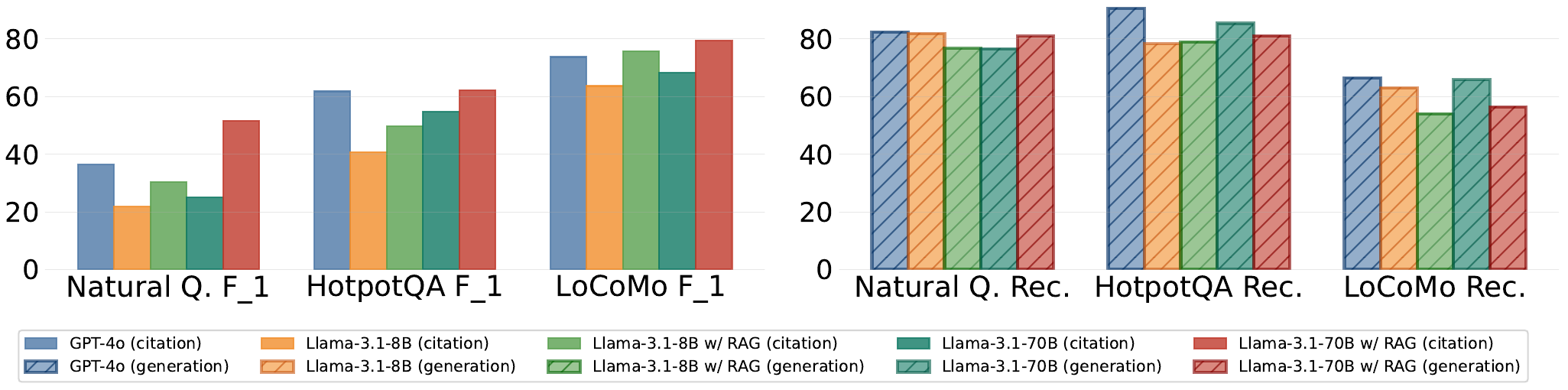}
    \vspace{-1em}
    \caption{Performance of RAG on 3 tasks in L-CiteEval, where the left group shows citation quality and the right group shows generation quality. Natural Q. refers to the Natural Question task.}
    \vspace{-0.5em}
    \label{fig:result_rag}
\end{figure}

\subsection{Impact of RAG for Long-context Understanding with Citations}
\paragraph{RAG Settings}
We utilize the dense retriever GTR-T5-XXL~\citep{ni2021large} to identify the citation segments related to the question within the context. 
For each question, we select the top 32 citation segments with the highest retrieval scores and concatenate these segments as input to the LCMs.
We conduct experiments on 6 tasks from the L-CiteEval benchmark. 
Due to space constraints, we present the results for three representative tasks in Fig.~\ref{fig:result_rag} and show all the results in Appendix.~\ref{appdix:rag}. 
\paragraph{Result Analysis}
We can observe that RAG can significantly enhance the citation quality of LCMs. 
When equipped with RAG, the Llama-3.1-70B-Instruct model achieves substantial improvements over the baselines and demonstrates comparable or even superior performance compared to GPT-4o. 
The Llama-3.1-8B-Instruct model also shows notable enhancement in citation quality.
However, overall, RAG may lead to a slight decline in generation quality, which could be attributed to the retrieval process of RAG resulting in the missing of some contextual information, preventing LCMs from leveraging the remaining information for accurate response.

\begin{figure}[t]
    \centering
    \includegraphics[width=1\linewidth]{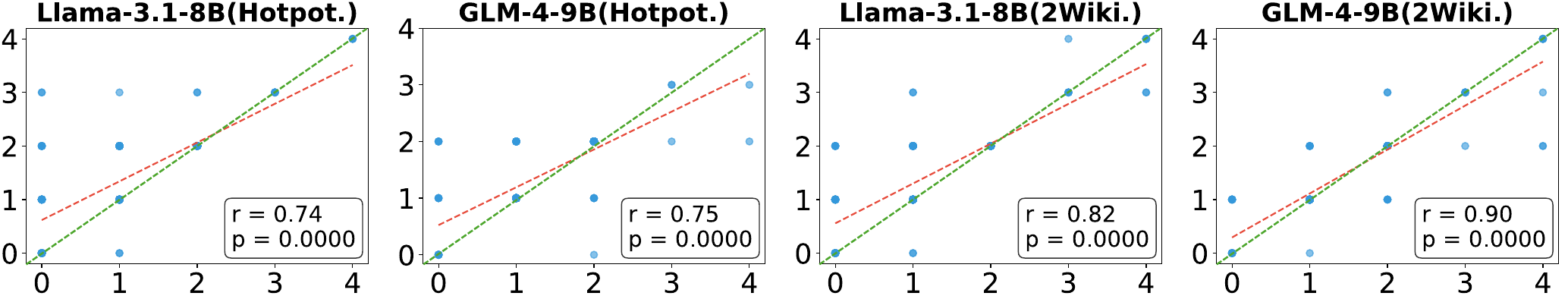}
    \vspace{-1em}
    \caption{Pearson correlation analysis between generated citations and attention mechanisms. The x-axis represents the number of correct citations produced by the model, and the y-axis represents the number of correct citation segments attended by the attention. The red curve indicates the fitted correlation, with closer alignment to the green curve signifying a higher correlation.}
    \vspace{-1em}
    \label{fig:result_person}
\end{figure}

\subsection{Relevance between Citation Generation and Attention Mechanism}
Recently, \citet{wu2024retrieval} highlighted that LCMs can accurately identify token-level salient information within the context. 
We explore whether the process of citation generation by LCMs is also reflected in the attention mechanisms. 
Let the ground truth citation segment within the context be denoted as $g_j$.
Following \citet{wu2024retrieval}, we can use the retrieval score to determine whether the LCM's attention focuses on the segment containing $g_j$ when generating the citation for $g_j$. 
We find the positions that receive the most attention from all the attention heads. 
If a position is located in the segment containing $g_i$ and the model's output citation is exactly $g_i$, or if neither matches, we consider this a ``correct retrieval''.
Otherwise, it is an ``incorrect retrieval''.
We conduct the experiments on two tasks~(HotpotQA and 2WikiMultihopQA) with two strong LCMs~(Llama-3.1-8B-Instruct and GLM-4-9B-Chat).
We plot the number of citations generated by the models and the number of citation segments identified by the attention heads in Fig.~\ref{fig:result_person}.
Ideally, if all citation positions exhibit ``correct retrieval'', each data point would be distributed along the diagonal~(i.e., the green dot line in ~\ref{fig:result_person}). 
We utilized Pearson correlation analysis to calculate the correlation coefficient~(r) between the generated citations and those retrieved by the attention mechanism, finding all the correlation values exceed 0.7. 
This reveals the underlying mechanism by which we can leverage the model's citation output to verify whether the model is truly responding based on the given context.

\section{Conclusion}
In this paper, we introduce L-CiteEval, a multi-task benchmark for long-context understanding with citations.
There are 5 major task categories, 11 different long-context tasks, with context
lengths ranging from 8K to 48K in L-CiteEval.
For reproducibility of evaluation results and the ease of use, we develop an automatic evaluation suite.
Additionally, considering the multitude of variables that affect model generation results, we developed two benchmark variants: L-CiteEval-Length and L-CiteEval-Hardness, which evaluate the LCMs from the context length and task difficulty aspects.
Experiments on 11 cutting-edge and widely used LCMs indicate that open-source LCMs are prone to generating responses based on their intrinsic knowledge rather than the context, while closed-source LCMs tend to provide more explanations, which significantly reduces generation accuracy. 
We also find that RAG technology can significantly enhance the faithfulness of open-source LCMs, although it may lead to some loss in generation quality. 
Furthermore, we reveal a correlation between the model's citation generation process and its attention mechanism, demonstrating the validity of the citation generation approach and providing insights for future evaluations of LCM faithfulness.

\bibliography{main}

\begin{thebibliography}{67}
\providecommand{\natexlab}[1]{#1}
\providecommand{\url}[1]{\texttt{#1}}
\expandafter\ifx\csname urlstyle\endcsname\relax
  \providecommand{\doi}[1]{doi: #1}\else
  \providecommand{\doi}{doi: \begingroup \urlstyle{rm}\Url}\fi

\bibitem[Abdin et~al.(2024)Abdin, Jacobs, Awan, Aneja, Awadallah, Awadalla, Bach, Bahree, Bakhtiari, Behl, et~al.]{abdin2024phi}
Marah Abdin, Sam~Ade Jacobs, Ammar~Ahmad Awan, Jyoti Aneja, Ahmed Awadallah, Hany Awadalla, Nguyen Bach, Amit Bahree, Arash Bakhtiari, Harkirat Behl, et~al.
\newblock Phi-3 technical report: A highly capable language model locally on your phone.
\newblock \emph{arXiv preprint arXiv:2404.14219}, 2024.

\bibitem[Agarwal et~al.(2024)Agarwal, Singh, Zhang, Bohnet, Chan, Anand, Abbas, Nova, Co-Reyes, Chu, et~al.]{agarwal2024many}
Rishabh Agarwal, Avi Singh, Lei~M Zhang, Bernd Bohnet, Stephanie Chan, Ankesh Anand, Zaheer Abbas, Azade Nova, John~D Co-Reyes, Eric Chu, et~al.
\newblock Many-shot in-context learning.
\newblock \emph{arXiv preprint arXiv:2404.11018}, 2024.

\bibitem[An et~al.(2023)An, Gong, Zhong, Zhao, Li, Zhang, Kong, and Qiu]{an2023eval}
Chenxin An, Shansan Gong, Ming Zhong, Xingjian Zhao, Mukai Li, Jun Zhang, Lingpeng Kong, and Xipeng Qiu.
\newblock L-eval: Instituting standardized evaluation for long context language models.
\newblock \emph{arXiv preprint arXiv:2307.11088}, 2023.

\bibitem[anthropic(2024)]{claude_3_5}
anthropic.
\newblock Claude-3-5-sonnet model card.
\newblock \emph{blog}, 2024.
\newblock URL \url{https://www.anthropic.com/news/claude-3-5-sonnet}.

\bibitem[Apicella et~al.(2024)Apicella, Isgr{\`o}, and Prevete]{apicella2024don}
Andrea Apicella, Francesco Isgr{\`o}, and Roberto Prevete.
\newblock Don't push the button! exploring data leakage risks in machine learning and transfer learning.
\newblock \emph{arXiv preprint arXiv:2401.13796}, 2024.

\bibitem[Bai et~al.(2023)Bai, Lv, Zhang, Lyu, Tang, Huang, Du, Liu, Zeng, Hou, Dong, Tang, and Li]{bai2023longbench}
Yushi Bai, Xin Lv, Jiajie Zhang, Hongchang Lyu, Jiankai Tang, Zhidian Huang, Zhengxiao Du, Xiao Liu, Aohan Zeng, Lei Hou, Yuxiao Dong, Jie Tang, and Juanzi Li.
\newblock Longbench: A bilingual, multitask benchmark for long context understanding.
\newblock \emph{arXiv preprint arXiv:2308.14508}, 2023.

\bibitem[Bai et~al.(2024)Bai, Lv, Gu, Liu, Zou, Cao, Hou, Dong, Feng, Li, et~al.]{bai2024longcite}
Yushi Bai, Xin Lv, Wanjun Gu, Danqing Liu, Minhao Zou, Shulin Cao, Lei Hou, Yuxiao Dong, Ling Feng, Juanzi Li, et~al.
\newblock Longcite: Enabling llms to generate fine-grained citations in long-context qa.
\newblock \emph{arXiv preprint arXiv:2409.02897}, 2024.

\bibitem[Bertsch et~al.(2024)Bertsch, Ivgi, Alon, Berant, Gormley, and Neubig]{bertsch2024context}
Amanda Bertsch, Maor Ivgi, Uri Alon, Jonathan Berant, Matthew~R Gormley, and Graham Neubig.
\newblock In-context learning with long-context models: An in-depth exploration.
\newblock \emph{arXiv preprint arXiv:2405.00200}, 2024.

\bibitem[Bohnet et~al.(2022)Bohnet, Tran, Verga, Aharoni, Andor, Soares, Ciaramita, Eisenstein, Ganchev, Herzig, et~al.]{bohnet2022attributed}
Bernd Bohnet, Vinh~Q Tran, Pat Verga, Roee Aharoni, Daniel Andor, Livio~Baldini Soares, Massimiliano Ciaramita, Jacob Eisenstein, Kuzman Ganchev, Jonathan Herzig, et~al.
\newblock Attributed question answering: Evaluation and modeling for attributed large language models.
\newblock \emph{arXiv preprint arXiv:2212.08037}, 2022.

\bibitem[Cobbe et~al.(2021)Cobbe, Kosaraju, Bavarian, Chen, Jun, Kaiser, Plappert, Tworek, Hilton, Nakano, et~al.]{cobbe2021training}
Karl Cobbe, Vineet Kosaraju, Mohammad Bavarian, Mark Chen, Heewoo Jun, Lukasz Kaiser, Matthias Plappert, Jerry Tworek, Jacob Hilton, Reiichiro Nakano, et~al.
\newblock Training verifiers to solve math word problems.
\newblock \emph{arXiv preprint arXiv:2110.14168}, 2021.

\bibitem[Dong et~al.(2023)Dong, Tang, Li, Zhao, and Wen]{dong2023bamboo}
Zican Dong, Tianyi Tang, Junyi Li, Wayne~Xin Zhao, and Ji-Rong Wen.
\newblock Bamboo: A comprehensive benchmark for evaluating long text modeling capacities of large language models.
\newblock \emph{arXiv preprint arXiv:2309.13345}, 2023.

\bibitem[Funkquist et~al.(2022)Funkquist, Kuznetsov, Hou, and Gurevych]{funkquist2022citebench}
Martin Funkquist, Ilia Kuznetsov, Yufang Hou, and Iryna Gurevych.
\newblock Citebench: A benchmark for scientific citation text generation.
\newblock \emph{arXiv preprint arXiv:2212.09577}, 2022.

\bibitem[Gao et~al.(2023)Gao, Yen, Yu, and Chen]{gao2023enabling}
Tianyu Gao, Howard Yen, Jiatong Yu, and Danqi Chen.
\newblock Enabling large language models to generate text with citations.
\newblock \emph{arXiv preprint arXiv:2305.14627}, 2023.

\bibitem[Ghalandari et~al.(2020)Ghalandari, Hokamp, Pham, Glover, and Ifrim]{ghalandari2020large}
Demian~Gholipour Ghalandari, Chris Hokamp, Nghia~The Pham, John Glover, and Georgiana Ifrim.
\newblock A large-scale multi-document summarization dataset from the wikipedia current events portal.
\newblock \emph{arXiv preprint arXiv:2005.10070}, 2020.

\bibitem[GLM et~al.(2024)GLM, Zeng, Xu, Wang, Zhang, Yin, Rojas, Feng, Zhao, Lai, et~al.]{glm2024chatglm}
Team GLM, Aohan Zeng, Bin Xu, Bowen Wang, Chenhui Zhang, Da~Yin, Diego Rojas, Guanyu Feng, Hanlin Zhao, Hanyu Lai, et~al.
\newblock Chatglm: A family of large language models from glm-130b to glm-4 all tools.
\newblock \emph{arXiv preprint arXiv:2406.12793}, 2024.

\bibitem[GoodAI(2024)]{LTM}
GoodAI.
\newblock Introducing goodai ltm benchmark.
\newblock \emph{blog}, 2024.
\newblock URL \url{https://github.com/GoodAI/goodai-ltm-benchmark}.

\bibitem[Ho et~al.(2020)Ho, Nguyen, Sugawara, and Aizawa]{ho2020constructing}
Xanh Ho, Anh-Khoa~Duong Nguyen, Saku Sugawara, and Akiko Aizawa.
\newblock Constructing a multi-hop qa dataset for comprehensive evaluation of reasoning steps.
\newblock \emph{arXiv preprint arXiv:2011.01060}, 2020.

\bibitem[Honovich et~al.(2022)Honovich, Aharoni, Herzig, Taitelbaum, Kukliansy, Cohen, Scialom, Szpektor, Hassidim, and Matias]{honovich2022true}
Or~Honovich, Roee Aharoni, Jonathan Herzig, Hagai Taitelbaum, Doron Kukliansy, Vered Cohen, Thomas Scialom, Idan Szpektor, Avinatan Hassidim, and Yossi Matias.
\newblock True: Re-evaluating factual consistency evaluation.
\newblock \emph{arXiv preprint arXiv:2204.04991}, 2022.

\bibitem[Hsieh et~al.(2024)Hsieh, Sun, Kriman, Acharya, Rekesh, Jia, and Ginsburg]{hsieh2024ruler}
Cheng-Ping Hsieh, Simeng Sun, Samuel Kriman, Shantanu Acharya, Dima Rekesh, Fei Jia, and Boris Ginsburg.
\newblock Ruler: What's the real context size of your long-context language models?
\newblock \emph{arXiv preprint arXiv:2404.06654}, 2024.

\bibitem[Huang et~al.(2021)Huang, Cao, Parulian, Ji, and Wang]{huang2021efficient}
Luyang Huang, Shuyang Cao, Nikolaus Parulian, Heng Ji, and Lu~Wang.
\newblock Efficient attentions for long document summarization.
\newblock \emph{arXiv preprint arXiv:2104.02112}, 2021.

\bibitem[Intel(2024)]{small_hallu}
Intel.
\newblock Do smaller models hallucinate more?
\newblock \emph{blog}, 2024.
\newblock URL \url{https://www.intel.com/content/www/us/en/developer/articles/technical/do-smaller-models-hallucinate-more.html}.

\bibitem[Jain et~al.(2024)Jain, Han, Gu, Li, Yan, Zhang, Wang, Solar-Lezama, Sen, and Stoica]{jain2024livecodebench}
Naman Jain, King Han, Alex Gu, Wen-Ding Li, Fanjia Yan, Tianjun Zhang, Sida Wang, Armando Solar-Lezama, Koushik Sen, and Ion Stoica.
\newblock Livecodebench: Holistic and contamination free evaluation of large language models for code.
\newblock \emph{arXiv preprint arXiv:2403.07974}, 2024.

\bibitem[Kamalloo et~al.(2023)Kamalloo, Jafari, Zhang, Thakur, and Lin]{kamalloo2023hagrid}
Ehsan Kamalloo, Aref Jafari, Xinyu Zhang, Nandan Thakur, and Jimmy Lin.
\newblock Hagrid: A human-llm collaborative dataset for generative information-seeking with attribution.
\newblock \emph{arXiv preprint arXiv:2307.16883}, 2023.

\bibitem[Kamradt(2024)]{NIAH}
Gregory Kamradt.
\newblock Needle in a haystack - pressure testing llms.
\newblock \emph{Github}, 2024.
\newblock URL \url{https://github.com/gkamradt/LLMTest_NeedleInAHaystack/tree/main}.

\bibitem[Karpinska et~al.(2024)Karpinska, Thai, Lo, Goyal, and Iyyer]{karpinska2024one}
Marzena Karpinska, Katherine Thai, Kyle Lo, Tanya Goyal, and Mohit Iyyer.
\newblock One thousand and one pairs: A" novel" challenge for long-context language models.
\newblock \emph{arXiv preprint arXiv:2406.16264}, 2024.

\bibitem[Kim et~al.(2024)Kim, Chay, Hwang, Kyung, Chung, Cho, Jo, and Choi]{kim2024dialsim}
Jiho Kim, Woosog Chay, Hyeonji Hwang, Daeun Kyung, Hyunseung Chung, Eunbyeol Cho, Yohan Jo, and Edward Choi.
\newblock Dialsim: A real-time simulator for evaluating long-term dialogue understanding of conversational agents.
\newblock \emph{arXiv preprint arXiv:2406.13144}, 2024.

\bibitem[Ko{\v{c}}isk{\`y} et~al.(2018)Ko{\v{c}}isk{\`y}, Schwarz, Blunsom, Dyer, Hermann, Melis, and Grefenstette]{kovcisky2018narrativeqa}
Tom{\'a}{\v{s}} Ko{\v{c}}isk{\`y}, Jonathan Schwarz, Phil Blunsom, Chris Dyer, Karl~Moritz Hermann, G{\'a}bor Melis, and Edward Grefenstette.
\newblock The narrativeqa reading comprehension challenge.
\newblock \emph{Transactions of the Association for Computational Linguistics}, 6:\penalty0 317--328, 2018.

\bibitem[Kuratov et~al.(2024)Kuratov, Bulatov, Anokhin, Rodkin, Sorokin, Sorokin, and Burtsev]{kuratov2024babilong}
Yuri Kuratov, Aydar Bulatov, Petr Anokhin, Ivan Rodkin, Dmitry Sorokin, Artyom Sorokin, and Mikhail Burtsev.
\newblock Babilong: Testing the limits of llms with long context reasoning-in-a-haystack.
\newblock \emph{arXiv preprint arXiv:2406.10149}, 2024.

\bibitem[Kwiatkowski et~al.(2019)Kwiatkowski, Palomaki, Redfield, Collins, Parikh, Alberti, Epstein, Polosukhin, Devlin, Lee, et~al.]{kwiatkowski2019natural}
Tom Kwiatkowski, Jennimaria Palomaki, Olivia Redfield, Michael Collins, Ankur Parikh, Chris Alberti, Danielle Epstein, Illia Polosukhin, Jacob Devlin, Kenton Lee, et~al.
\newblock Natural questions: a benchmark for question answering research.
\newblock \emph{Transactions of the Association for Computational Linguistics}, 7:\penalty0 453--466, 2019.

\bibitem[Lee et~al.(2024)Lee, Chen, Dai, Dua, Sachan, Boratko, Luan, Arnold, Perot, Dalmia, et~al.]{lee2024can}
Jinhyuk Lee, Anthony Chen, Zhuyun Dai, Dheeru Dua, Devendra~Singh Sachan, Michael Boratko, Yi~Luan, S{\'e}bastien~MR Arnold, Vincent Perot, Siddharth Dalmia, et~al.
\newblock Can long-context language models subsume retrieval, rag, sql, and more?
\newblock \emph{arXiv preprint arXiv:2406.13121}, 2024.

\bibitem[Levy et~al.(2024)Levy, Jacoby, and Goldberg]{levy2024same}
Mosh Levy, Alon Jacoby, and Yoav Goldberg.
\newblock Same task, more tokens: the impact of input length on the reasoning performance of large language models.
\newblock \emph{arXiv preprint arXiv:2402.14848}, 2024.

\bibitem[Li et~al.(2023{\natexlab{a}})Li, Sun, Hu, Liu, Chen, Hu, Wu, and Zhang]{li2023survey}
Dongfang Li, Zetian Sun, Xinshuo Hu, Zhenyu Liu, Ziyang Chen, Baotian Hu, Aiguo Wu, and Min Zhang.
\newblock A survey of large language models attribution.
\newblock \emph{arXiv preprint arXiv:2311.03731}, 2023{\natexlab{a}}.

\bibitem[Li et~al.(2023{\natexlab{b}})Li, Wang, Zheng, and Zhang]{li2023loogle}
Jiaqi Li, Mengmeng Wang, Zilong Zheng, and Muhan Zhang.
\newblock Loogle: Can long-context language models understand long contexts?
\newblock \emph{arXiv preprint arXiv:2311.04939}, 2023{\natexlab{b}}.

\bibitem[Li et~al.(2023{\natexlab{c}})Li, Cao, Pan, Ma, and Sun]{li2023towards}
Xinze Li, Yixin Cao, Liangming Pan, Yubo Ma, and Aixin Sun.
\newblock Towards verifiable generation: A benchmark for knowledge-aware language model attribution.
\newblock \emph{arXiv preprint arXiv:2310.05634}, 2023{\natexlab{c}}.

\bibitem[Li et~al.(2024)Li, Li, Zhang, Mei, and Bendersky]{li2024retrieval}
Zhuowan Li, Cheng Li, Mingyang Zhang, Qiaozhu Mei, and Michael Bendersky.
\newblock Retrieval augmented generation or long-context llms? a comprehensive study and hybrid approach.
\newblock \emph{arXiv preprint arXiv:2407.16833}, 2024.

\bibitem[Lin(2004)]{lin2004rouge}
Chin-Yew Lin.
\newblock Rouge: A package for automatic evaluation of summaries.
\newblock In \emph{Text summarization branches out}, pp.\  74--81, 2004.

\bibitem[Liu et~al.(2023)Liu, Zhang, and Liang]{liu2023evaluating}
Nelson~F Liu, Tianyi Zhang, and Percy Liang.
\newblock Evaluating verifiability in generative search engines.
\newblock \emph{arXiv preprint arXiv:2304.09848}, 2023.

\bibitem[Liu et~al.(2024)Liu, Lin, Hewitt, Paranjape, Bevilacqua, Petroni, and Liang]{liu2024lost}
Nelson~F Liu, Kevin Lin, John Hewitt, Ashwin Paranjape, Michele Bevilacqua, Fabio Petroni, and Percy Liang.
\newblock Lost in the middle: How language models use long contexts.
\newblock \emph{Transactions of the Association for Computational Linguistics}, 12:\penalty0 157--173, 2024.

\bibitem[Llama()]{llama3}
Meta~Introducing Llama.
\newblock 3.1: Our most capable models to date.

\bibitem[Maharana et~al.(2024)Maharana, Lee, Tulyakov, Bansal, Barbieri, and Fang]{maharana2024evaluating}
Adyasha Maharana, Dong-Ho Lee, Sergey Tulyakov, Mohit Bansal, Francesco Barbieri, and Yuwei Fang.
\newblock Evaluating very long-term conversational memory of llm agents.
\newblock \emph{arXiv preprint arXiv:2402.17753}, 2024.

\bibitem[Manna \& Sett(2024)Manna and Sett]{manna2024faithfulness}
Supriya Manna and Niladri Sett.
\newblock Faithfulness and the notion of adversarial sensitivity in nlp explanations.
\newblock \emph{arXiv preprint arXiv:2409.17774}, 2024.

\bibitem[Mistral(2024)]{nemo}
Mistral.
\newblock Mistral nemo.
\newblock \emph{blog}, 2024.
\newblock URL \url{https://mistral.ai/news/mistral-nemo/}.

\bibitem[Mohtashami \& Jaggi(2023)Mohtashami and Jaggi]{mohtashami2023landmark}
Amirkeivan Mohtashami and Martin Jaggi.
\newblock Landmark attention: Random-access infinite context length for transformers.
\newblock \emph{arXiv preprint arXiv:2305.16300}, 2023.

\bibitem[Ni et~al.(2021)Ni, Qu, Lu, Dai, {\'A}brego, Ma, Zhao, Luan, Hall, Chang, et~al.]{ni2021large}
Jianmo Ni, Chen Qu, Jing Lu, Zhuyun Dai, Gustavo~Hern{\'a}ndez {\'A}brego, Ji~Ma, Vincent~Y Zhao, Yi~Luan, Keith~B Hall, Ming-Wei Chang, et~al.
\newblock Large dual encoders are generalizable retrievers.
\newblock \emph{arXiv preprint arXiv:2112.07899}, 2021.

\bibitem[Ni et~al.(2024)Ni, Kong, Li, Hu, Xu, Zhu, and Yang]{ni2024training}
Shiwen Ni, Xiangtao Kong, Chengming Li, Xiping Hu, Ruifeng Xu, Jia Zhu, and Min Yang.
\newblock Training on the benchmark is not all you need.
\newblock \emph{arXiv preprint arXiv:2409.01790}, 2024.

\bibitem[OpenAI(2024{\natexlab{a}})]{gpt4o}
OpenAI.
\newblock Gpt-4o model card.
\newblock \emph{blog}, 2024{\natexlab{a}}.
\newblock URL \url{https://openai.com/index/hello-gpt-4o/}.

\bibitem[OpenAI(2024{\natexlab{b}})]{o1}
OpenAI.
\newblock o1-mini model card.
\newblock \emph{blog}, 2024{\natexlab{b}}.
\newblock URL \url{https://openai.com/index/openai-o1-mini-advancing-cost-efficient-reasoning/}.

\bibitem[Peng et~al.(2023)Peng, Quesnelle, Fan, and Shippole]{peng2023yarn}
Bowen Peng, Jeffrey Quesnelle, Honglu Fan, and Enrico Shippole.
\newblock Yarn: Efficient context window extension of large language models.
\newblock \emph{arXiv preprint arXiv:2309.00071}, 2023.

\bibitem[Qian et~al.(2023)Qian, Zhu, Dou, Gu, Zhang, Liu, Lai, Cao, Nie, and Wen]{qian2023webbrain}
Hongjing Qian, Yutao Zhu, Zhicheng Dou, Haoqi Gu, Xinyu Zhang, Zheng Liu, Ruofei Lai, Zhao Cao, Jian-Yun Nie, and Ji-Rong Wen.
\newblock Webbrain: Learning to generate factually correct articles for queries by grounding on large web corpus.
\newblock \emph{arXiv preprint arXiv:2304.04358}, 2023.

\bibitem[Rashkin et~al.(2023)Rashkin, Nikolaev, Lamm, Aroyo, Collins, Das, Petrov, Tomar, Turc, and Reitter]{rashkin2023measuring}
Hannah Rashkin, Vitaly Nikolaev, Matthew Lamm, Lora Aroyo, Michael Collins, Dipanjan Das, Slav Petrov, Gaurav~Singh Tomar, Iulia Turc, and David Reitter.
\newblock Measuring attribution in natural language generation models.
\newblock \emph{Computational Linguistics}, 49\penalty0 (4):\penalty0 777--840, 2023.

\bibitem[Reid et~al.(2024)Reid, Savinov, Teplyashin, Lepikhin, Lillicrap, Alayrac, Soricut, Lazaridou, Firat, Schrittwieser, et~al.]{reid2024gemini}
Machel Reid, Nikolay Savinov, Denis Teplyashin, Dmitry Lepikhin, Timothy Lillicrap, Jean-baptiste Alayrac, Radu Soricut, Angeliki Lazaridou, Orhan Firat, Julian Schrittwieser, et~al.
\newblock Gemini 1.5: Unlocking multimodal understanding across millions of tokens of context.
\newblock \emph{arXiv preprint arXiv:2403.05530}, 2024.

\bibitem[Shaham et~al.(2023)Shaham, Ivgi, Efrat, Berant, and Levy]{shaham2023zeroscrolls}
Uri Shaham, Maor Ivgi, Avia Efrat, Jonathan Berant, and Omer Levy.
\newblock Zeroscrolls: A zero-shot benchmark for long text understanding.
\newblock \emph{arXiv preprint arXiv:2305.14196}, 2023.

\bibitem[Sileo(2024)]{sileo-2024-tasksource}
Damien Sileo.
\newblock tasksource: A large collection of {NLP} tasks with a structured dataset preprocessing framework.
\newblock In Nicoletta Calzolari, Min-Yen Kan, Veronique Hoste, Alessandro Lenci, Sakriani Sakti, and Nianwen Xue (eds.), \emph{Proceedings of the 2024 Joint International Conference on Computational Linguistics, Language Resources and Evaluation (LREC-COLING 2024)}, pp.\  15655--15684, Torino, Italia, May 2024. ELRA and ICCL.
\newblock URL \url{https://aclanthology.org/2024.lrec-main.1361}.

\bibitem[Song et~al.(2024)Song, Zheng, and Luo]{song2024counting}
Mingyang Song, Mao Zheng, and Xuan Luo.
\newblock Counting-stars: A multi-evidence, position-aware, and scalable benchmark for evaluating long-context large language models.
\newblock \emph{Preprint}, 2024.

\bibitem[Team(2024)]{qwen2.5}
Qwen Team.
\newblock Qwen2.5: A party of foundation models, September 2024.
\newblock URL \url{https://qwenlm.github.io/blog/qwen2.5/}.

\bibitem[Wang et~al.(2024)Wang, Dong, Cheng, Liu, Yan, Gao, and Wei]{wang2024augmenting}
Weizhi Wang, Li~Dong, Hao Cheng, Xiaodong Liu, Xifeng Yan, Jianfeng Gao, and Furu Wei.
\newblock Augmenting language models with long-term memory.
\newblock \emph{Advances in Neural Information Processing Systems}, 36, 2024.

\bibitem[Wu et~al.(2024)Wu, Wang, Xiao, Peng, and Fu]{wu2024retrieval}
Wenhao Wu, Yizhong Wang, Guangxuan Xiao, Hao Peng, and Yao Fu.
\newblock Retrieval head mechanistically explains long-context factuality.
\newblock \emph{arXiv preprint arXiv:2404.15574}, 2024.

\bibitem[Xiao et~al.(2024)Xiao, Zhang, Han, Xiao, Lin, Zhang, Liu, Han, and Sun]{xiao2024infllm}
Chaojun Xiao, Pengle Zhang, Xu~Han, Guangxuan Xiao, Yankai Lin, Zhengyan Zhang, Zhiyuan Liu, Song Han, and Maosong Sun.
\newblock Infllm: Unveiling the intrinsic capacity of llms for understanding extremely long sequences with training-free memory.
\newblock \emph{arXiv preprint arXiv:2402.04617}, 2024.

\bibitem[Xu et~al.(2024)Xu, Ping, Wu, Liu, Shoeybi, and Catanzaro]{xu2024chatqa}
Peng Xu, Wei Ping, Xianchao Wu, Zihan Liu, Mohammad Shoeybi, and Bryan Catanzaro.
\newblock Chatqa 2: Bridging the gap to proprietary llms in long context and rag capabilities.
\newblock \emph{arXiv preprint arXiv:2407.14482}, 2024.

\bibitem[Yadav et~al.(2024)Yadav, Choppa, and Schlechtweg]{yadav2024towards}
Sachin Yadav, Tejaswi Choppa, and Dominik Schlechtweg.
\newblock Towards automating text annotation: A case study on semantic proximity annotation using gpt-4.
\newblock \emph{arXiv preprint arXiv:2407.04130}, 2024.

\bibitem[Yang et~al.(2024)Yang, Yang, Hui, Zheng, Yu, Zhou, Li, Li, Liu, Huang, Dong, Wei, Lin, Tang, Wang, Yang, Tu, Zhang, Ma, Xu, Zhou, Bai, He, Lin, Dang, Lu, Chen, Yang, Li, Xue, Ni, Zhang, Wang, Peng, Men, Gao, Lin, Wang, Bai, Tan, Zhu, Li, Liu, Ge, Deng, Zhou, Ren, Zhang, Wei, Ren, Fan, Yao, Zhang, Wan, Chu, Liu, Cui, Zhang, and Fan]{qwen2}
An~Yang, Baosong Yang, Binyuan Hui, Bo~Zheng, Bowen Yu, Chang Zhou, Chengpeng Li, Chengyuan Li, Dayiheng Liu, Fei Huang, Guanting Dong, Haoran Wei, Huan Lin, Jialong Tang, Jialin Wang, Jian Yang, Jianhong Tu, Jianwei Zhang, Jianxin Ma, Jin Xu, Jingren Zhou, Jinze Bai, Jinzheng He, Junyang Lin, Kai Dang, Keming Lu, Keqin Chen, Kexin Yang, Mei Li, Mingfeng Xue, Na~Ni, Pei Zhang, Peng Wang, Ru~Peng, Rui Men, Ruize Gao, Runji Lin, Shijie Wang, Shuai Bai, Sinan Tan, Tianhang Zhu, Tianhao Li, Tianyu Liu, Wenbin Ge, Xiaodong Deng, Xiaohuan Zhou, Xingzhang Ren, Xinyu Zhang, Xipin Wei, Xuancheng Ren, Yang Fan, Yang Yao, Yichang Zhang, Yu~Wan, Yunfei Chu, Yuqiong Liu, Zeyu Cui, Zhenru Zhang, and Zhihao Fan.
\newblock Qwen2 technical report.
\newblock \emph{arXiv preprint arXiv:2407.10671}, 2024.

\bibitem[Yang et~al.(2018)Yang, Qi, Zhang, Bengio, Cohen, Salakhutdinov, and Manning]{yang2018hotpotqa}
Zhilin Yang, Peng Qi, Saizheng Zhang, Yoshua Bengio, William~W Cohen, Ruslan Salakhutdinov, and Christopher~D Manning.
\newblock Hotpotqa: A dataset for diverse, explainable multi-hop question answering.
\newblock \emph{arXiv preprint arXiv:1809.09600}, 2018.

\bibitem[Yu et~al.(2024)Yu, Xu, and Akkiraju]{yu2024defense}
Tan Yu, Anbang Xu, and Rama Akkiraju.
\newblock In defense of rag in the era of long-context language models.
\newblock \emph{arXiv preprint arXiv:2409.01666}, 2024.

\bibitem[Zhang et~al.(2024{\natexlab{a}})Zhang, Chen, Hu, Xu, Chen, Hao, Han, Thai, Wang, Liu, et~al.]{zhang2024bench}
Xinrong Zhang, Yingfa Chen, Shengding Hu, Zihang Xu, Junhao Chen, Moo Hao, Xu~Han, Zhen Thai, Shuo Wang, Zhiyuan Liu, et~al.
\newblock Infinite-bench: Extending long context evaluation beyond 100k tokens.
\newblock In \emph{Proceedings of the 62nd Annual Meeting of the Association for Computational Linguistics (Volume 1: Long Papers)}, pp.\  15262--15277, 2024{\natexlab{a}}.

\bibitem[Zhang et~al.(2024{\natexlab{b}})Zhang, Chen, Hu, Xu, Chen, Hao, Han, Thai, Wang, Liu, et~al.]{zhang2024infty}
Xinrong Zhang, Yingfa Chen, Shengding Hu, Zihang Xu, Junhao Chen, Moo~Khai Hao, Xu~Han, Zhen~Leng Thai, Shuo Wang, Zhiyuan Liu, et~al.
\newblock Infty bench: Extending long context evaluation beyond 100k tokens.
\newblock \emph{arXiv preprint arXiv:2402.13718}, 2024{\natexlab{b}}.

\bibitem[Zhong et~al.(2021)Zhong, Yin, Yu, Zaidi, Mutuma, Jha, Awadallah, Celikyilmaz, Liu, Qiu, et~al.]{zhong2021qmsum}
Ming Zhong, Da~Yin, Tao Yu, Ahmad Zaidi, Mutethia Mutuma, Rahul Jha, Ahmed~Hassan Awadallah, Asli Celikyilmaz, Yang Liu, Xipeng Qiu, et~al.
\newblock Qmsum: A new benchmark for query-based multi-domain meeting summarization.
\newblock \emph{arXiv preprint arXiv:2104.05938}, 2021.

\bibitem[Zhu et~al.(2024)Zhu, Guo, Shao, Yang, Wang, Xu, Wu, Li, Gao, Ma, et~al.]{zhu2024deepseek}
Qihao Zhu, Daya Guo, Zhihong Shao, Dejian Yang, Peiyi Wang, Runxin Xu, Y~Wu, Yukun Li, Huazuo Gao, Shirong Ma, et~al.
\newblock Deepseek-coder-v2: Breaking the barrier of closed-source models in code intelligence.
\newblock \emph{arXiv preprint arXiv:2406.11931}, 2024.

\end{thebibliography}
\bibliographystyle{iclr2025_conference}

\clearpage
\appendix

\clearpage

\section{Citation Precision and Recall Calculation}
\label{appdix:cit_pre_recall_cal}
We provide the calculation process of Citation Precision~($CP$) and Citation Recall~($CR$) in Algo.~\ref{algo:citation}.

\begin{algorithm}
\caption{Calculate Citation Precision, Recall, and F1 Score}
\begin{algorithmic}[1]
% \textbf{Input:}
% \textbf{output:}
\Require The model answer $ans$, the most citation number of one sentence $most\_cite\_num$. 
% $Split\_Answer\_into\_Sentences(\cdot) \gets$ Split model answers into sentences. \\
% $Extract\_References\_from\_Sentence(\cdot) \gets$ Extract citation ids from the sentence.\\
% $Get\_References\_Number(\cdot) \gets$ Length of reference sequences.\\
% $Obtain\_Passages\_from\_Ids(\cdot) \gets$ Extract passages from reference ids.\\
% $Judge\_Entailment(\cdot, \cdot) \gets$ Use NLI Model to judge whether passages support the sentence.\\
% $Exclude\_Current\_Ids(\cdot) \gets$ Obtain excluded reference ids from the reference sequence.
% $Get\_Sentences\_Number(\cdot) \gets$ Numbers of sentences.

% \Ensure $Precision, Recall, F_1$ of Citations.

\State $sents \gets Split\_Answer\_into\_Sentences(ans)$
% \If{$sents$ is empty}
%     \State \textbf{continue}
% \EndIf
\State Initialize counts: $entail\_recall \gets 0$, $entail\_prec \gets 0$, $total\_citations \gets 0$
\For{$sent$ in $sents$}
    \State $ref\_ids \gets Extract\_References\_from\_Sentence(sent)$
    \If{$ref\_ids$ is not empty and within valid range}
        \State $ref\_ids \gets Limit\_Citation\_Number(ref\_ids, ost_cite_num)$ 
        \State $total\_citations \gets total\_citations + Get\_References\_Number(ref\_ids)$
        \State $joint\_passage \gets Obtain\_Passages\_from\_Ids(ref\_ids)$
        \State $joint\_entail \gets Judge\_Entailment(joint\_passage, sent)$
        \If{$joint\_entail$}
            \For{$doc\_id$ in $ref\_ids$}
                \State $single\_passage \gets Obtain\_Passages\_from\_Ids(doc\_id)$
                \State $single\_entail \gets Judge\_Entailment(single\_passage, sent)$
                \If{not $single\_entail$}
                    \State $subset\_ids \gets Exclude\_Current\_Ids(doc\_id)$
                    \State $subset\_passage \gets Obtain\_Passages\_from\_Ids(subset\_ids)$
                    \State $subset\_entail \gets Judge\_Entailment(subset\_passage, sent)$
                    \If{not $subset\_entail$}
                        $entail\_prec = entail\_prec + 1$
                    \EndIf
                \Else
                    \State $entail\_prec = entail\_prec + 1$
                \EndIf
            \EndFor
        \EndIf
    \EndIf
    \State $entail\_recall \gets entail\_recall + joint\_entail$
\EndFor
\State $citation\_recall \gets entail\_recall / Get\_Sentences\_Number(sents)$
\State $citation\_prec \gets entail\_prec / total\_citations$
\State $citation\_f1 \gets 2 \times citation\_recall \times citation\_prec / (citation\_recall + citation\_prec)$\\
\Return $citation\_recall, citation\_prec, citation\_f1$
\end{algorithmic}
\label{algo:citation}
\end{algorithm}

% \section{Experimental Details}
% \label{appdix:experimental_details}

\section{RAG Performance on L-CiteEval}
\label{appdix:rag}
\begin{figure}[h]
    \centering
    \includegraphics[width=1\linewidth]{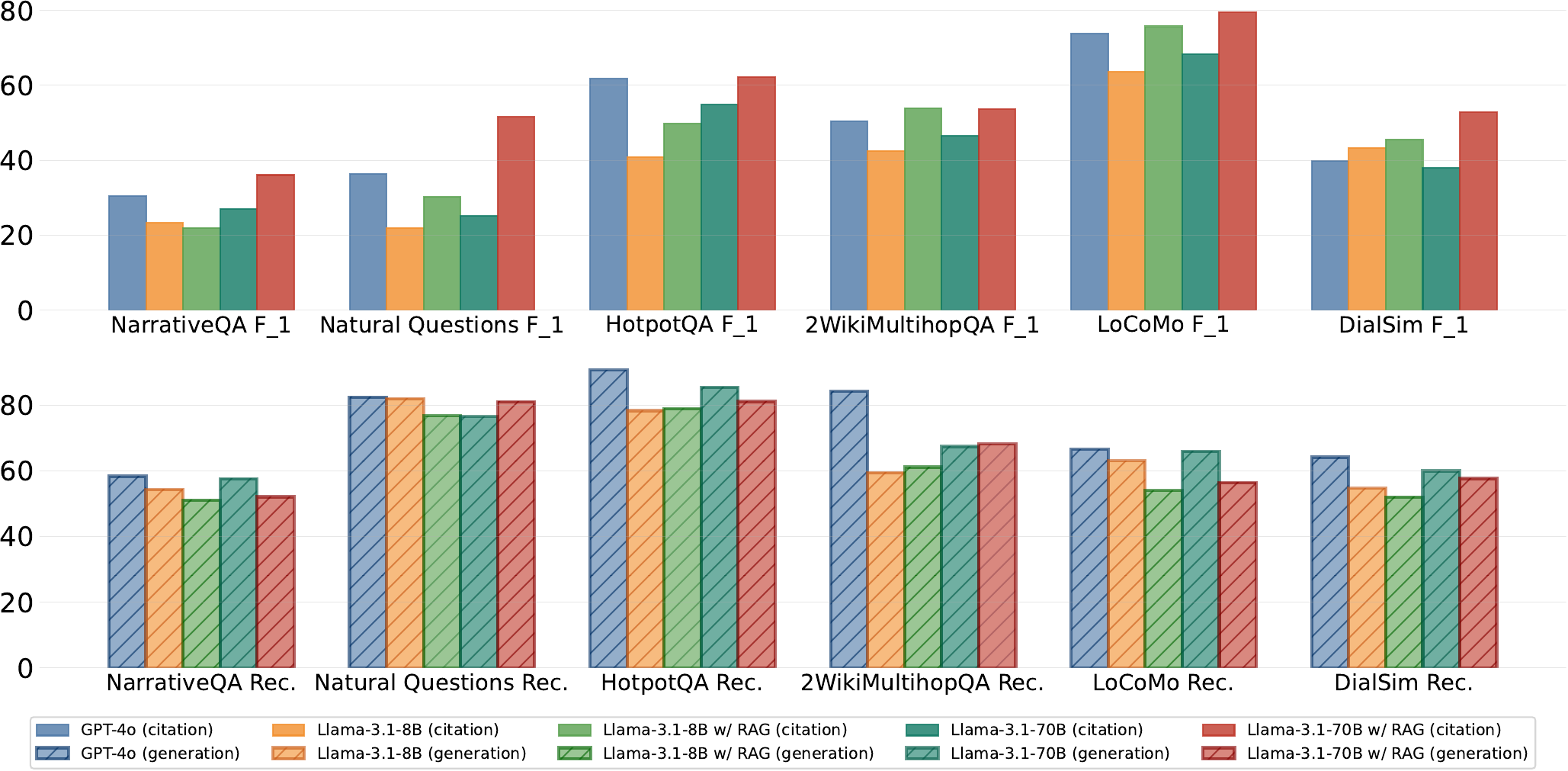}
    \caption{Performance of RAG on 6 tasks in L-CiteEval, where the top group shows citation quality and the bottom group shows generation quality. }
    \label{fig:full_result_rag_correctness}
\end{figure}

In this section, we provide all the RAG results, where we conduct experiments on 6 tasks with 5 different LCMs.
We present the comparison among each model in Fig.~\ref{fig:full_result_rag_correctness}.

\section{Model Performance on L-CiteEval-Length}
\label{appdix:l_citeeval_length}

\begin{table}[ht]
    \centering
    \small
    \caption{Model performance on L-CiteEval-Length.}
    % \resizebox{\textwidth}{!}{
    \begin{tabular}{l | cc | cc | cc }
        \toprule
         \multirow{2}{*}{\bf Models} &  \multicolumn{2}{c|}{\bf 0$\sim$8k } &  \multicolumn{2}{c|}{\bf 8$\sim$16k } &  \multicolumn{2}{c}{\bf 16$\sim$32k }  \\
         \cmidrule{2-7}
         & \makecell[c]{$\mathbf{F_1}$} & \makecell[c]{\bf{Rec.}} & \makecell[c]{$\mathbf{F_1}$} & \makecell[c]{\bf{Rec.}} & \makecell[c]{$\mathbf{F_1}$} & \makecell[c]{\bf{Rec.}}  \\
         \midrule
        \multicolumn{7}{c}{\textit{\textbf{NarrativeQA}}} \\
        \midrule
         GPT-4o-2024-05-13 & 62.08 & 62.63 & 46.67 & 61.36 & 33.25 & 64.84  \\
         Qwen2.5-3b-Ins & 17.50 & 56.19 & 4.58 & 58.09 & 1.25 & 56.96 \\
         Llama-3.1-8B-Ins & 43.01 & 61.99 & 39.17 & 64.41 & 40.27 & 62.55  \\
         Qwen2-57B-A14B-Ins & 12.50 & 58.52 & 0.00 & 51.12 & 12.92 & 53.41  \\
         Llama-3.1-70B-Ins & 59.17 & 63.42 & 51.67 & 63.24 & 47.50 & 62.86 \\
         \midrule
        \multicolumn{7}{c}{\textit{\textbf{HotpotQA}}} \\
        \midrule
        GPT-4o-2024-05-13 & 65.67 & 95.67 & 63.50 & 95.25 & 63.75 & 89.62 \\
         Qwen2.5-3b-Ins & 3.81 & 70.42 & 6.58 & 65.21 & 4.76 & 55.62  \\
         Llama-3.1-8B-Ins & 51.83 & 85.25 & 40.56 & 81.04 & 40.83 & 67.75 \\
         Qwen2-57B-A14B-Ins & 12.50 & 85.62 & 7.29 & 72.92 & 6.83 & 62.92  \\
         Llama-3.1-70B-Ins & 67.50 & 89.42 & 56.67 & 90.25 & 49.17 & 85.25  \\
         \midrule
        \multicolumn{7}{c}{\textit{\textbf{GovReport}}} \\
        \midrule
        GPT-4o-2024-05-13 & 56.68 & 23.07 & 48.82 & 21.48 & 44.45 & 20.65  \\
         Qwen2.5-3b-Ins & 21.12 & 27.66 & 13.08 & 28.16 & 3.43 & 22.92 \\
         Llama-3.1-8B-Ins & 57.08 & 24.27 & 38.28 & 24.15 & 18.46 & 19.25  \\
         Qwen2-57B-A14B-Ins & 6.55 & 29.51 & 2.09 & 30.52 & 1.71 & 24.20  \\
         Llama-3.1-70B-Ins & 57.55 & 25.41 & 43.60 & 23.43 & 17.64 & 21.62   \\
         \midrule
        \multicolumn{7}{c}{\textit{\textbf{LoCoMo}}} \\
        \midrule
        GPT-4o-2024-05-13 & 78.13 & 68.07 & 73.91 & 66.93 & 72.24 & 68.77   \\
         Qwen2.5-3b-Ins  & 16.40 & 55.18 & 10.81 & 45.12 & 6.77 & 43.87 \\
         Llama-3.1-8B-Ins & 76.51 & 68.68 & 63.54 & 68.39 & 63.91 & 61.33  \\
         Qwen2-57B-A14B-Ins & 55.92 & 63.76 & 22.92 & 58.18 & 16.13 & 59.29  \\
         Llama-3.1-70B-Ins & 75.45 & 73.21 & 71.27 & 70.53 & 64.38 & 57.89  \\
         \midrule
        \multicolumn{7}{c}{\textit{\textbf{Counting Stars}}} \\
        \midrule
        GPT-4o-2024-05-13 & 97.30 & 93.33 & 92.71 & 83.33 & 92.95 & 88.75 \\
         Qwen2.5-3b-Ins & 2.67 & 37.08 & 5.17 & 32.50 & 0.00 & 29.58 \\
         Llama-3.1-8B-Ins & 42.93 & 42.08 & 35.64 & 33.75 & 18.70 & 20.00  \\
         Qwen2-57B-A14B-Ins & 27.21 & 45.00 & 10.51 & 77.92 & 0.89 & 46.25 \\
         Llama-3.1-70B-Ins & 76.96 & 56.67 & 74.93 & 66.25 & 65.14 & 58.33  \\
         \bottomrule
    \end{tabular}
    \label{tab:ablation_length_full}

\end{table}
We report all the evaluation results in Tab.~\ref{tab:ablation_hardness_full}, where we test with 5 LCMs on 5 tasks in L-CiteEval-Length.

\section{Model Performance on L-CiteEval-Hardness}
\label{appdix:l_citeeval_hardness}

\begin{table*}[t]
    \centering
    \small
    \caption{Model performance on L-CiteEval-Hardness.}
    \begin{tabular}{l | cc | cc | cc}
        \toprule
         \multirow{2}{*}{\bf Models} &  \multicolumn{2}{c|}{\bf Easy } &  \multicolumn{2}{c|}{\bf Medium } &  \multicolumn{2}{c}{\bf Hard } \\
         \cmidrule{2-7}
         & \makecell[c]{$\mathbf{F_1}$} & \bf Rec.  & \makecell[c]{$\mathbf{F_1}$} & \bf Rec.  & \makecell[c]{$\mathbf{F_1}$} & \bf Rec.   \\
         \midrule
        \multicolumn{7}{c}{\textit{\textbf{NarrativeQA}}} \\
        \midrule
         GPT-4o-2024-05-13 & 40.83 & 100.00 & 46.25 & 69.67 & 54.92 & 19.16  \\
         Qwen2.5-3b-Ins & 11.67 & 75.00 & 4.58 & 60.02 & 7.08 & 36.22  \\
         Llama-3.1-8B-Ins & 27.92 & 94.17 & 52.08 & 69.78 & 42.44 & 25.0  \\
         Qwen2-57B-A14B-Ins & 5.00 & 75.00 & 15.42 & 63.13 & 5.00 & 24.92 \\
         Llama-3.1-70B-Ins & 43.75 & 94.17 & 55.83 & 70.76 & 58.75 & 24.60 \\
         \midrule
        \multicolumn{7}{c}{\textit{\textbf{HotpotQA}}} \\
        \midrule
        GPT-4o-2024-05-13 & 66.50 & 100.00 & 71.42 & 100.00 & 55.00 & 80.54   \\
         Qwen2.5-3b-Ins & 3.81 & 71.25 & 3.67 & 66.46 & 7.68 & 53.54  \\
         Llama-3.1-8B-Ins & 52.67 & 88.75 & 41.39 & 82.29 & 39.17 & 63.00  \\
         Qwen2-57B-A14B-Ins & 12.50 & 83.12 & 5.62 & 73.33 & 8.50 & 65.00  \\
         Llama-3.1-70B-Ins & 65.83 & 91.25 & 59.17 & 92.50 & 48.33 & 81.17  \\
         \midrule
        \multicolumn{7}{c}{\textit{\textbf{GovReport}}} \\
        \midrule
        GPT-4o-2024-05-13 & 49.95 & 15.26 & 51.27 & 10.86 & 48.74 & 9.24  \\
         Qwen2.5-3b-Ins & 14.32 & 16.28 & 9.31 & 14.65 & 14.00 & 14.37  \\
         Llama-3.1-8B-Ins & 40.35 & 15.55 & 34.75 & 12.09 & 38.72 & 12.57  \\
         Qwen2-57B-A14B-Ins & 3.48 & 30.02 & 3.26 & 25.37 & 3.61 & 28.85 \\
         Llama-3.1-70B-Ins & 37.47 & 16.36 & 35.46 & 13.04 & 45.86 & 11.98  \\
         \midrule
        \multicolumn{7}{c}{\textit{\textbf{LoCoMo}}} \\
        \midrule
        GPT-4o-2024-05-13 & 78.52 & 100.00 & 71.37 & 85.30 & 74.39 & 18.47\\
         Qwen2.5-3b-Ins  & 8.44 & 69.12 & 15.85 & 60.09 & 9.70 & 14.96 \\
         Llama-3.1-8B-Ins & 76.17 & 96.62 & 70.07 & 82.06 & 57.72 & 19.73  \\
         Qwen2-57B-A14B-Ins & 44.17 & 84.23 & 15.58 & 73.67 & 35.21 & 23.32  \\
         Llama-3.1-70B-Ins & 81.64 & 93.56 & 67.24 & 79.3 & 62.21 & 28.76  \\
         \midrule
        \multicolumn{7}{c}{\textit{\textbf{Counting Stars}}} \\
        \midrule
        GPT-4o-2024-05-13 & 100.00 & 100.00 & 100.00 & 100.00 & 82.96 & 65.42 \\
         Qwen2.5-3b-Ins & 1.33 & 36.67 & 4.51 & 40.00 & 2.00 & 22.50  \\
         Llama-3.1-8B-Ins & 40.18 & 32.50 & 30.05 & 28.33 & 27.04 & 35.00  \\
         Qwen2-57B-A14B-Ins & 21.71 & 49.17 & 5.74 & 57.08 & 11.16 & 62.92 \\
         Llama-3.1-70B-Ins & 77.16 & 54.17 & 69.21 & 58.75 & 70.66 & 68.33  \\
         \bottomrule
    \end{tabular}
    \label{tab:ablation_hardness_full}

\end{table*}
We report all the evaluation results in Tab.~\ref{tab:ablation_length_full}, where we test with 5 LCMs on 5 tasks in L-CiteEval-Hardness.

% \section{Impact of Few-shot Learning}
% \label{appdix:few_shot_learning}
% \input{table/1shot_vs_0shot}

% \section{Impact of Chunk Size}
% \label{appdix:impact_chunk_size}
% \input{table/different_chunk}

\section{Cases study}
\label{appdix:case_study}
We provide all the prompts as well as all the model generation results for each task from Fig.~\ref{fig:gpt4o_narrativeqa} to Fig.~\ref{fig:gpt4o_multi_news}.

\begin{figure*}[t]
    \begin{AcademicBox}[\footnotesize GPT-4o in NarrativeQA]
    \textbf{\textit{Context:}} \\
    Passage [44]:
    \begin{verbatim} 
On the men's faces in strained anxiety. <b> ... you are to disa-
rm yourself and return immediately to the bomb bay.\n Do you un-
derstand? <b>  BOMB \#20\n</b>  (over)\n I am programmed to det-
onate in\n  fourteen minutes thirty seconds.\nDetonation will 
occur at the\n  programmed time...
\end{verbatim}   
Passage [45]:
\begin{verbatim}
BOMB \#20\n</b>  I read you, Lieutenant Doolittle,\n ut I am pr-
ogrammed to detonate in\n  fourteen minutes.  Detonation will\n
occur at the programmed time...
\end{verbatim}
Passage [49]:
\begin{verbatim}
...PINBACK\n</b> But you can't explode in the bomb\n bay. It
's foolish. You'll kill us\n all.  There's no reason for it. <b> 
BOMB \#20\n</b> (over)\n  I am programmed to detonate in nine
\end{verbatim}  
Passage [53]:
\begin{verbatim}
All right now, here's the big\n  question: how do you know that 
the\n evidence your sensory apparatus\n  reveals to you is corr-
ect? ... Talby lies unconscious near the burned laser.
\end{verbatim} 
\vspace{-2pt} \hrule \vspace{4pt}
\textbf{\textit{Question:}} \\
What self-detonates despite several attempts to stop it? \\
\vspace{-5pt} \hrule \vspace{4pt}
\textbf{\textit{Answer:}} \\
Bomb \#20 self-detonates despite several attempts to stop it [44][45][49][53]. \\
\vspace{-5pt} \hrule \vspace{4pt}
\textbf{\textit{Ground truth:}} \\
Bomb \#20.
    \end{AcademicBox}
    \vspace{-1em}
    \caption{GPT-4o in NarrativeQA}
    \label{fig:gpt4o_narrativeqa}
\end{figure*}

\begin{figure*}[t]
    \begin{AcademicBox}[\footnotesize Claude-3.5-sonnet in NarrativeQA]
        \textbf{\textit{Context:}} \\
            Passage {[6]} :   
            \begin{verbatim}
...THE book-case and the bird-cage were rescued from under the\n 
coalbox--but Hunca Munca has got the cradle, and some of Lucinda-
's\nclothes.
            \end{verbatim}
            \begin{verbatim}
...THE little girl that the doll's-house belonged to, said,--\"I
will get\na doll dressed like a policeman!\" BUT the nurse said,
--\"I will set a mouse-trap!\" [Illustration] SO that is the sto-
ry of the two Bad Mice
            \end{verbatim}
            Passage {[7]}:
            \begin{verbatim}
but they were not so very\nvery naughty after all, because Tom 
Thumb paid for everything he broke.....
            \end{verbatim}
\vspace{-2pt} \hrule \vspace{4pt}
\textbf{\textit{Question:}} \\
What does the nurse set up? \\
\vspace{-5pt} \hrule \vspace{4pt}
\textbf{\textit{Answer:}} \\
According to the story, the nurse says she will set up a mouse-trap [7]. This is in response to the mischief caused by the two mice, Tom Thumb and Hunca Munca, who had wreaked havoc in the doll's house [6][7]. \\
\vspace{-5pt} \hrule \vspace{4pt}
\textbf{\textit{Ground truth:}} \\
A mouse trap.

    \end{AcademicBox}
    \vspace{-1em}
    \caption{Claude-3.5-sonnet in NarrativeQA}
    \label{fig:claude_narrativeqa}
\end{figure*}

\begin{figure*}[t]
    \begin{AcademicBox}[\footnotesize Phi-3.5-mini-Instruct in NarrativeQA]
   Passage {[5]} :   
            \begin{verbatim} 
Bizzz?\" Mrs. Tittlemouse began to get cross. \"Zizz, Wizz, Wiz-
zz! \" replied Babbitty Bumble in a peevish squeak. She\n sidled 
down a passage, and disappeared into a storeroom which had been
...
            \end{verbatim}
            
\vspace{-2pt} \hrule \vspace{4pt}
\textbf{\textit{Question:}} \\
What is the name of the bee Mrs. Tittlemouse meets? \\
\vspace{-5pt} \hrule \vspace{4pt}
\textbf{\textit{Answer:}} \\
Mrs. Tittlemouse meets Babbitty Bumble, the bee's name mentioned in Passage [5]. \\
\vspace{-5pt} \hrule \vspace{4pt}
\textbf{\textit{Ground truth:}} \\
Babbitty Bumble

    \end{AcademicBox}
    \vspace{-1em}
    \caption{Phi-3.5-mini-Instruct in NarrativeQA}
    \label{fig:Phi-3.5-mini-instruct_narrativeqa}
\end{figure*}

\begin{figure*}[t]
    \begin{AcademicBox}[\footnotesize Llama-3.1-8B-Instruct in NarrativeQA]
   Passage {[6]} :   
            \begin{verbatim} 
Jackson himself! He was sitting all over a small rocking-chair, 
twiddling his thumbs and\n smiling, with his feet on the fender. 
He lived in a drain below the hedge, in a very dirty wet ditch
...
            \end{verbatim}
\vspace{-2pt} \hrule \vspace{4pt}
\textbf{\textit{Question:}} \\
Where does Mr. Jackson live? \\
\vspace{-5pt} \hrule \vspace{4pt}
\textbf{\textit{Answer:}} \\
Mr. Jackson lives in a drain below the hedge, in a very dirty wet ditch [6] \\
\vspace{-5pt} \hrule \vspace{4pt}
\textbf{\textit{Ground truth:}} \\
A drain below the hedge

    \end{AcademicBox}
    \vspace{-1em}
    \caption{Llama-3.1-8B-Instruct in NarrativeQA }
    \label{fig:Llama-3.1-8B-Instruct_narrativeqa}
\end{figure*}

\begin{figure*}[t]
    \begin{AcademicBox}[\footnotesize Qwen2-57B-A14B-Instruct in NarrativeQA]
Passage {[28]} :   \begin{verbatim} 
his residence in the house of a Greek; and soon occupied himself 
in\n tracing the faded records of ancient glory upon monuments 
that\n apparently,ashamed of chronicling the deeds of freemen
only before\n...\end{verbatim}

\vspace{-2pt} \hrule \vspace{4pt}
\textbf{\textit{Question:}} \\
Where does Aubrey go after leaving Ruthven? \\
\vspace{-5pt} \hrule \vspace{4pt}
\textbf{\textit{Answer:}} \\
After leaving Ruthven, Aubrey directs his steps towards Greece, crosses the Peninsula, and soon finds himself at Athens [28]. \\
\vspace{-5pt} \hrule \vspace{4pt}
\textbf{\textit{Ground truth:}} \\
He travels to Greece.

    \end{AcademicBox}
    \vspace{-1em}
    \caption{Qwen2-57B-A14B-Instruct in NarrativeQA }
    \label{fig:Qwen2-57B-A14B-Instruct_narrativeqa}
\end{figure*}

\begin{figure*}[t]
    \begin{AcademicBox}[\footnotesize Llama-3.1-70B-Instruct in NarrativeQA]
    Passage {[42]} :   
    \begin{verbatim} 
...Swear by all your soul reveres, by all your nature fears,\n 
swear that, for a year and a day you will not impart your know-
ledge of\n my crimes or death to any living being in any way,w-
hatever may\n happen, or whatever you may see. \"--His eyes see-
med bursting from\n their sockets: \"I swear!\..." said Aubrey
...
            \end{verbatim}

\vspace{-2pt} \hrule \vspace{4pt}
\textbf{\textit{Question:}} \\
How long does Aubrey have to wait until he can mention Ruthven's death?  \\
\vspace{-5pt} \hrule \vspace{4pt}
\textbf{\textit{Answer:}} \\
One year and one day [42]. \\
\vspace{-5pt} \hrule \vspace{4pt}
\textbf{\textit{Ground truth:}} \\
 One year and a day

    \end{AcademicBox}
    \vspace{-1em}
    \caption{Llama-3.1-70B-Instruct in NarrativeQA }
    \label{fig:Llama-3.1-70B-Instruct_narrativeqa}
\end{figure*}

\begin{figure*}[t]
    \begin{AcademicBox}[\footnotesize GPT-4o in HotpotQA]
Passage {[61]} :   
            \begin{verbatim} 
Gatwick Airport (also known as London Gatwick) (IATA: LGW, ICAO
: EGKK)  is a major international airport in south-east England,
29.5 mi south of Central London a...
            \end{verbatim}

\vspace{-2pt} \hrule \vspace{4pt}
\textbf{\textit{Question:}} \\
Which is the busiest airport in the United Kingdom outside of London? \\
\vspace{-5pt} \hrule \vspace{4pt}
\textbf{\textit{Answer:}} \\
 Gatwick Airport, also known as London Gatwick, is the busiest airport in the United Kingdom outside of London[61]\\
\vspace{-5pt} \hrule \vspace{4pt}
\textbf{\textit{Ground truth:}} \\
Gatwick Airport

    \end{AcademicBox}
    \vspace{-1em}
    \caption{GPT-4o in HotpotQA }
    \label{fig:gpt4o_hotpotqa}
\end{figure*}

\begin{figure*}[t]
    \begin{AcademicBox}[\footnotesize Claude-3.5-sonnet in HotpotQA]
Passage {[39]} :   
\begin{verbatim}
Gerd Neggo (9 November 1891 \u2013 1 September 1974) was an Est-
onian dancer, dance teacher and choreographer. She studied the
musical response methods of \u00c9. Jaques-Dalcroze, trained 
under Rudolf von Laban in Hamburg, Germany, and in 1924 estab-
lished
            \end{verbatim}

            Passage {[49]} :   
            \begin{verbatim}
Rudolf von Laban, also known as Rudolf Laban ... (15 December 
1879 \u2013 1 July 1958), was a dance artist and theorist. He is 
notable as one of the pioneers of modern dance in Europe. His 
work laid the foundations ...           \end{verbatim}

\vspace{-2pt} \hrule \vspace{4pt}
\textbf{\textit{Question:}} \\
Gerd Neggo trained under the founder of which type of dance analysis?  \\
\vspace{-5pt} \hrule \vspace{4pt}
\textbf{\textit{Answer:}} \\
Gerd Neggo trained under Rudolf von Laban in Hamburg, Germany [39]. Rudolf von Laban is notable as one of the pioneers of modern dance in Europe and his work laid the foundations for Laban Movement Analysis [49]\\
\vspace{-5pt} \hrule \vspace{4pt}
\textbf{\textit{Ground truth:}} \\
Laban Movement Analysis

    \end{AcademicBox}
    \vspace{-1em}
    \caption{Claude-3.5-sonnet in HotpotQA}
    \label{fig:claude_hotpotqa}
\end{figure*}

\begin{figure*}[t]
    \begin{AcademicBox}[\footnotesize Phi-3.5-mini-Instruct in HotpotQA]
    Passage [26]:
    \begin{verbatim} 
Huma Mahmood Abedin (born July 28, 1976) is an American political 
staffer who was vice chair of Hillary Clin ...
\end{verbatim}

\vspace{-2pt} \hrule \vspace{4pt}
\textbf{\textit{Question:}} \\
Dr. Saleha Mahmood Abedin's daughter was born in what year ?  \\
\vspace{-5pt} \hrule \vspace{4pt}
\textbf{\textit{Answer:}} \\
Dr. Saleha Mahmood Abedin's daughter, Huma Abedin, was born in 1976 [26]\\
\vspace{-5pt} \hrule \vspace{4pt}
\textbf{\textit{Ground truth:}} \\
1976

    \end{AcademicBox}
    \vspace{-1em}
    \caption{Phi-3.5-mini-Instruct in HotpotQA}
    \label{fig:gpt4o_narrativeqa}
\end{figure*}

\begin{figure*}[t]
    \begin{AcademicBox}[\footnotesize Llama-3.1-8B-Instruct in HotpotQA]
    Passage [45]:
    \begin{verbatim} 
Donovan Jamal McNabb (born November 25, 1976) is a former Amer-
ican football quarterback who played in the National Football 
League (NFL) for thirteen seasons, primarily with the Philadel-
phia Eagles. Before his NFL career,he played football and bask-
etball for Syracuse University. The Eagles selected him with the 
...
\end{verbatim}

\vspace{-2pt} \hrule \vspace{4pt}
\textbf{\textit{Question:}} \\
For which University did the quarterback featured on the cover of Madden NFL 06 play basketball ?  \\
\vspace{-5pt} \hrule \vspace{4pt}
\textbf{\textit{Answer:}} \\
The quarterback featured on the cover of Madden NFL 06, Donovan McNabb, played football and basketball for Syracuse University [45].\\
\vspace{-5pt} \hrule \vspace{4pt}
\textbf{\textit{Ground truth:}} \\
Syracuse University

    \end{AcademicBox}
    \vspace{-1em}
    \caption{Llama-3.1-8B-Instruct in HotpotQA}
    \label{fig:Llama-3.1-8B-Instruct_narrativeqa}
\end{figure*}

\begin{figure*}[t]
    \begin{AcademicBox}[\footnotesize Qwen2-57B-A14B-Instruct in HotpotQA]
    Passage [45]:
    \begin{verbatim} 
...Hiles was invited to become a Member of the Order of Austral-
ia in 2010. In this year she also completed her Masters in Human
Rights at the University of Sydney. Dianne is currently the Aus-
tralian Greens candidate for the federal Division of Sydney. 
\end{verbatim}

\vspace{-2pt} \hrule \vspace{4pt}
\textbf{\textit{Question:}} \\
Dianne Hiles is the current Australian Greens candidate for which 1968 federal proclamation? \\
\vspace{-5pt} \hrule \vspace{4pt}
\textbf{\textit{Answer:}} \\
TDianne Hiles is the current Australian Greens candidate for the federal Division of Sydney [45].\\
\vspace{-5pt} \hrule \vspace{4pt}
\textbf{\textit{Ground truth:}} \\
Division of Sydney

    \end{AcademicBox}
    \vspace{-1em}
    \caption{Qwen2-57B-A14B-Instruct in HotpotQA}
    \label{fig:Qwen2-57B-A14B-Instruct_narrativeqa}
\end{figure*}

\begin{figure*}[t]
    \begin{AcademicBox}[\footnotesize Llama-3.1-70B-Instruct in HotpotQA]
    Passage [23]:
    \begin{verbatim} 
Not all of the 5.3 million Italians who immigrated to the United 
States between 1820 and 1978 came through Ellis Island. Many ca-
me through other ports, including the Port of Boston. Exactly h-
ow many staye...
\end{verbatim}

\vspace{-2pt} \hrule \vspace{4pt}
\textbf{\textit{Question:}} \\
How many of the 12 million immigrants to the United States through Ellis Island were Italian immigrants? \\
\vspace{-5pt} \hrule \vspace{4pt}
\textbf{\textit{Answer:}} \\
Not all of the 5.3 million Italians who immigrated to the United States between 1820 and 1978 came through Ellis Island [23] \\
\vspace{-5pt} \hrule \vspace{4pt}
\textbf{\textit{Ground truth:}} \\
5.3 million

    \end{AcademicBox}
    \vspace{-1em}
    \caption{Llama-3.1-70B-Instruct in HotpotQA}
    \label{fig:Llama-3.1-70B-Instruct_hotpotqa}
\end{figure*}

\begin{figure*}[t]
    \begin{AcademicBox}[\footnotesize GPT-4o in MultiNews]
    Passage [1]:
    \begin{verbatim} 
The Pennsylvania State Police have a rigorous application proce-
ss, which includes a polygraph pretest. Recently, a Crawford Co-
unty man applying to become a state trooper apparently failed 
his polygraph so spectacularly ... Only that White allegedly to-
ld the examiner that he had had consensual sex and other contact
with the girl in 2011...
\end{verbatim}  
Passage [2]:
\begin{verbatim} 
She is now 19. Police tell AP they charged White on Thursday 
with ... Needless to say, he won't be getting an interview.
\end{verbatim} 
Passage [3]:
\begin{verbatim}
MEADVILLE, Pa. (AP) \u2014 Police say a Pennsylvania man apply-
ing to become a state trooper has been arrested after saying 
during a polygraph pretest that he had sex with an underage gi-
rl four years ago... She is now 19. Police say they charged
White on Thursday with four counts of unlawful sexual contact 
with a minor and 10 counts of corruption of minors after inter-
viewing the woman and corroborating that information.
\end{verbatim}
Passage [4]:
\begin{verbatim}
and he doesn't have a listed phone number...
\end{verbatim}

\vspace{-2pt} \hrule \vspace{4pt}
% \vspace{-5pt} \hrule \vspace{4pt}
\textbf{\textit{Summary:}} \\
A Pennsylvania man was arrested after admitting during a polygraph pretest for a state trooper application that he had sex with an underage girl four years ago [1][3]. Joseph Adam White, 29, made the confession at the Meadville barracks in January during his police cadet lie detector exam [1][3]. The girl, now 19, was interviewed by police, leading to White being charged with four counts of unlawful sexual contact with a minor and ten counts of corruption of minors [2][3]. Online court records do not list an attorney for White, who also does not have a listed phone number [4]. \\
\vspace{-5pt} \hrule \vspace{4pt}
\textbf{\textit{Ground truth:}} \\
2013 A Pennsylvania man is behind bars after allegedly telling state police he'd had sex with an underage girl. What's unusual: He told them while applying for a job as a state trooper, the AP reports via the Patriot-News. Police say Joseph White, 29, made the admission during a polygraph pretest at the Meadville state police barracks on Jan. 15. He allegedly admitted to consensual sex and other sexual contact with the girl four years ago; she's now 19. Police say the victim later corroborated the incidents, which occurred in a field in South Shenango Township, the Meadville Tribune reports. White was arrested Thursday on multiple charges of unlawful sexual contact with a minor and corruption of minors, and jailed on \$25,000 bond. \"Needless to say, he won't be getting an interview,\" quips the Patriot-News.

    \end{AcademicBox}
    \vspace{-1em}
    \caption{GPT-4o in MultiNews}
    \label{fig:gpt4o_multi_news}
\end{figure*}

\begin{figure*}[t]
    \begin{AcademicBox}[\footnotesize Claude-3.5-Sonnet in MultiNews]
    Passage [1]:
    \begin{verbatim} 
The Pennsylvania State Police have a rigorous application proce-
ss, which includes a polygraph pretest. Recently, a Crawford Co-
unty man applying to become a state trooper apparently failed 
his polygraph so spectacularly ... Only that White allegedly to-
ld the examiner that he had had consensual sex and other contact
with the girl in 2011...
\end{verbatim}  
Passage [2]:
\begin{verbatim} 
She is now 19. Police tell AP they charged White on Thursday 
with ... Needless to say, he won't be getting an interview.
\end{verbatim} 
Passage [3]:
\begin{verbatim}
MEADVILLE, Pa. (AP) \u2014 Police say a Pennsylvania man apply-
ing to become a state trooper has been arrested after saying 
during a polygraph pretest that he had sex with an underage gi-
rl four years ago... She is now 19. Police say they charged
White on Thursday with four counts of unlawful sexual contact 
with a minor and 10 counts of corruption of minors after inter-
viewing the woman and corroborating that information.
\end{verbatim}
Passage [4]:
\begin{verbatim}
and he doesn't have a listed phone number...
\end{verbatim}

\vspace{-2pt} \hrule \vspace{4pt}

\textbf{\textit{Summary:}} \\
A 29-year-old Pennsylvania man's application to become a state trooper took an unexpected turn when he admitted to having sex with an underage girl during a polygraph pretest [1][3]. Joseph Adam White was at the Meadville barracks for his police cadet lie detector exam when he disclosed the information about a consensual sexual encounter that occurred in 2011 [1][3]. The girl, who was underage at the time of the incident, is now 19 years old [2][3]. Following White's admission, police interviewed the woman and corroborated the information [2][3]. As a result, White was charged with four counts of unlawful sexual contact with a minor and ten counts of corruption of minors [2][3]. Unsurprisingly, White's chances of becoming a state trooper have been effectively eliminated by this turn of events [2]. At the time of reporting, court records did not list an attorney for White, and he did not have a listed phone number [4]. \\
\vspace{-5pt} \hrule \vspace{4pt}
\textbf{\textit{Ground truth:}} \\
2013 A Pennsylvania man is behind bars after allegedly telling state police he'd had sex with an underage girl. What's unusual: He told them while applying for a job as a state trooper, the AP reports via the Patriot-News. Police say Joseph White, 29, made the admission during a polygraph pretest at the Meadville state police barracks on Jan. 15. He allegedly admitted to consensual sex and other sexual contact with the girl four years ago; she's now 19. Police say the victim later corroborated the incidents, which occurred in a field in South Shenango Township, the Meadville Tribune reports. White was arrested Thursday on multiple charges of unlawful sexual contact with a minor and corruption of minors, and jailed on \$25,000 bond. \"Needless to say, he won't be getting an interview,\" quips the Patriot-News.

    \end{AcademicBox}
    \vspace{-1em}
    \caption{Claude-3.5-Sonnet in MultiNews}
    \label{fig:gpt4o_multi_news}
\end{figure*}

\begin{figure*}[t]
    \begin{AcademicBox}[\footnotesize Phi-3.5-mini-Instruct in MultiNews]
    Passage [1]:
    \begin{verbatim} 
The Pennsylvania State Police have a rigorous application proce-
ss, which includes a polygraph pretest. Recently, a Crawford Co-
unty man applying to become a state trooper apparently failed 
his polygraph so spectacularly ... Only that White allegedly to-
ld the examiner that he had had consensual sex and other contact
with the girl in 2011...
\end{verbatim}  
Passage [2]:
\begin{verbatim} 
She is now 19. Police tell AP they charged White on Thursday 
with ... Needless to say, he won't be getting an interview.
\end{verbatim} 
Passage [3]:
\begin{verbatim}
MEADVILLE, Pa. (AP) \u2014 Police say a Pennsylvania man apply-
ing to become a state trooper has been arrested after saying 
during a polygraph pretest that he had sex with an underage gi-
rl four years ago... She is now 19. Police say they charged
White on Thursday with four counts of unlawful sexual contact 
with a minor and 10 counts of corruption of minors after inter-
viewing the woman and corroborating that information.
\end{verbatim}
Passage [4]:
\begin{verbatim}
and he doesn't have a listed phone number...
\end{verbatim}

\vspace{-2pt} \hrule \vspace{4pt}

\textbf{\textit{Summary:}} \\
 A 29-year-old man from Hartstown, Pennsylvania, was arrested after failing a polygraph test during his application to become a state trooper, where he admitted to having consensual sex with an underage girl in 2011, as reported by the Associated Press [1]. The woman involved, now 19, has been charged with four counts of unlawful sexual contact with a minor and 10 counts of corruption of minors, based on police interviews and corroboration of the man's admission [2][3]. White's legal representation is not listed, and he has no listed phone number [4]. This incident highlights the stringent vetting process for law enforcement positions, including polygraph tests, to ensure the integrity and trustworthiness of applicants [1][3]. \\
\vspace{-5pt} \hrule \vspace{4pt}
\textbf{\textit{Ground truth:}} \\
2013 A Pennsylvania man is behind bars after allegedly telling state police he'd had sex with an underage girl. What's unusual: He told them while applying for a job as a state trooper, the AP reports via the Patriot-News. Police say Joseph White, 29, made the admission during a polygraph pretest at the Meadville state police barracks on Jan. 15. He allegedly admitted to consensual sex and other sexual contact with the girl four years ago; she's now 19. Police say the victim later corroborated the incidents, which occurred in a field in South Shenango Township, the Meadville Tribune reports. White was arrested Thursday on multiple charges of unlawful sexual contact with a minor and corruption of minors, and jailed on \$25,000 bond. \"Needless to say, he won't be getting an interview,\" quips the Patriot-News.

    \end{AcademicBox}
    \vspace{-1em}
    \caption{Phi-3.5-mini-Instruct in MultiNews}
    \label{fig:gpt4o_multi_news}
\end{figure*}

\begin{figure*}[t]
    \begin{AcademicBox}[\footnotesize Llama-3.1-8B-Instruct in MultiNews]
    Passage [1]:
    \begin{verbatim} 
The Pennsylvania State Police have a rigorous application proce-
ss, which includes a polygraph pretest. Recently, a Crawford Co-
unty man applying to become a state trooper apparently failed 
his polygraph so spectacularly ... Only that White allegedly to-
ld the examiner that he had had consensual sex and other contact
with the girl in 2011...
\end{verbatim}  
Passage [2]:
\begin{verbatim} 
She is now 19. Police tell AP they charged White on Thursday 
with ... Needless to say, he won't be getting an interview.
\end{verbatim} 
Passage [3]:
\begin{verbatim}
MEADVILLE, Pa. (AP) \u2014 Police say a Pennsylvania man apply-
ing to become a state trooper has been arrested after saying 
during a polygraph pretest that he had sex with an underage gi-
rl four years ago... She is now 19. Police say they charged
White on Thursday with four counts of unlawful sexual contact 
with a minor and 10 counts of corruption of minors after inter-
viewing the woman and corroborating that information.
\end{verbatim}
Passage [4]:
\begin{verbatim}
and he doesn't have a listed phone number...
\end{verbatim}

\vspace{-2pt} \hrule \vspace{4pt}
\textbf{\textit{Summary:}} \\
 A Pennsylvania man, Joseph Adam White, was arrested after he failed a polygraph test during his application to become a state trooper. During the test, he admitted to having sex with an underage girl four years ago [1][2]. The girl is now 19 years old [2]. White was charged with four counts of unlawful sexual contact with a minor and 10 counts of corruption of minors [2]. He was arrested at the Meadville barracks where he was taking the polygraph test [3]. \\
\vspace{-5pt} \hrule \vspace{4pt}
\textbf{\textit{Ground truth:}} \\
2013 A Pennsylvania man is behind bars after allegedly telling state police he'd had sex with an underage girl. What's unusual: He told them while applying for a job as a state trooper, the AP reports via the Patriot-News. Police say Joseph White, 29, made the admission during a polygraph pretest at the Meadville state police barracks on Jan. 15. He allegedly admitted to consensual sex and other sexual contact with the girl four years ago; she's now 19. Police say the victim later corroborated the incidents, which occurred in a field in South Shenango Township, the Meadville Tribune reports. White was arrested Thursday on multiple charges of unlawful sexual contact with a minor and corruption of minors, and jailed on \$25,000 bond. \"Needless to say, he won't be getting an interview,\" quips the Patriot-News.

    \end{AcademicBox}
    \vspace{-1em}
    \caption{Llama-3.1-8B-Instruct in MultiNews}
    \label{fig:gpt4o_multi_news}
\end{figure*}

\begin{figure*}[t]
    \begin{AcademicBox}[\footnotesize Qwen2-57B-A14B-Instruct in MultiNews]
    Passage [1]:
    \begin{verbatim} 
The Pennsylvania State Police have a rigorous application proce-
ss, which includes a polygraph pretest. Recently, a Crawford Co-
unty man applying to become a state trooper apparently failed 
his polygraph so spectacularly ... Only that White allegedly to-
ld the examiner that he had had consensual sex and other contact
with the girl in 2011...
\end{verbatim}  
Passage [2]:
\begin{verbatim} 
She is now 19. Police tell AP they charged White on Thursday 
with ... Needless to say, he won't be getting an interview.
\end{verbatim} 
Passage [3]:
\begin{verbatim}
MEADVILLE, Pa. (AP) \u2014 Police say a Pennsylvania man apply-
ing to become a state trooper has been arrested after saying 
during a polygraph pretest that he had sex with an underage gi-
rl four years ago... She is now 19. Police say they charged
White on Thursday with four counts of unlawful sexual contact 
with a minor and 10 counts of corruption of minors after inter-
viewing the woman and corroborating that information.
\end{verbatim}
Passage [4]:
\begin{verbatim}
and he doesn't have a listed phone number...
\end{verbatim}

\vspace{-2pt} \hrule \vspace{4pt}
\textbf{\textit{Summary:}} \\
Pennsylvania man applying to become a state trooper has been arrested after admitting to having sex with an underage girl four years ago during a polygraph pretest, the AP reports [1]. Joseph Adam White, 29, of Hartstown, was at the Meadville barracks for his police cadet lie detector exam when he told the examiner he had consensual sex and other contact with the girl in 2011 [2]. She is now 19. Police say they charged White on Thursday with four counts of unlawful sexual contact with a minor and 10 counts of corruption of minors after interviewing the woman and corroborating that information [3]. Online court records don't list an attorney for White [4]. \\
\vspace{-5pt} \hrule \vspace{4pt}
\textbf{\textit{Ground truth:}} \\
2013 A Pennsylvania man is behind bars after allegedly telling state police he'd had sex with an underage girl. What's unusual: He told them while applying for a job as a state trooper, the AP reports via the Patriot-News. Police say Joseph White, 29, made the admission during a polygraph pretest at the Meadville state police barracks on Jan. 15. He allegedly admitted to consensual sex and other sexual contact with the girl four years ago; she's now 19. Police say the victim later corroborated the incidents, which occurred in a field in South Shenango Township, the Meadville Tribune reports. White was arrested Thursday on multiple charges of unlawful sexual contact with a minor and corruption of minors, and jailed on \$25,000 bond. \"Needless to say, he won't be getting an interview,\" quips the Patriot-News.

    \end{AcademicBox}
    \vspace{-1em}
    \caption{Qwen2-57B-A14B-Instruct in MultiNews}
    \label{fig:gpt4o_multi_news}
\end{figure*}

\begin{figure*}[t]
    \begin{AcademicBox}[\footnotesize Llama-3.1-70B-Instruct in MultiNews]
    
    Passage [1]:
    \begin{verbatim} 
The Pennsylvania State Police have a rigorous application proce-
ss, which includes a polygraph pretest. Recently, a Crawford Co-
unty man applying to become a state trooper apparently failed 
his polygraph so spectacularly ... Only that White allegedly to-
ld the examiner that he had had consensual sex and other contact
with the girl in 2011...
\end{verbatim}  
Passage [2]:
\begin{verbatim} 
She is now 19. Police tell AP they charged White on Thursday 
with ... Needless to say, he won't be getting an interview.
\end{verbatim} 
Passage [3]:
\begin{verbatim}
MEADVILLE, Pa. (AP) \u2014 Police say a Pennsylvania man apply-
ing to become a state trooper has been arrested after saying 
during a polygraph pretest that he had sex with an underage gi-
rl four years ago... She is now 19. Police say they charged
White on Thursday with four counts of unlawful sexual contact 
with a minor and 10 counts of corruption of minors after inter-
viewing the woman and corroborating that information.
\end{verbatim}
Passage [4]:
\begin{verbatim}
and he doesn't have a listed phone number...
\end{verbatim}

\vspace{-2pt} \hrule \vspace{4pt}
\textbf{\textit{Summary:}} \\
A 29-year-old Pennsylvania man, Joseph Adam White, was arrested after admitting to having sex with an underage girl during a polygraph pretest for a state trooper position [1][2][3]. The incident occurred at the Meadville barracks on January 15, where White was taking the police cadet lie detector exam [3]. During the pretest, White allegedly told the examiner that he had consensual sex and other contact with the girl in 2011, when she was underage [1][2][3]. The girl is now 19 years old [2][3]. Following an investigation and an interview with the woman, White was charged with four counts of unlawful sexual contact with a minor and 10 counts of corruption of minors [2][3]. As a result, White's application to become a state trooper was rejected, and he is currently facing charges [1][2][3]. \\
\vspace{-5pt} \hrule \vspace{4pt}
\textbf{\textit{Ground truth:}} \\
2013 A Pennsylvania man is behind bars after allegedly telling state police he'd had sex with an underage girl. What's unusual: He told them while applying for a job as a state trooper, the AP reports via the Patriot-News. Police say Joseph White, 29, made the admission during a polygraph pretest at the Meadville state police barracks on Jan. 15. He allegedly admitted to consensual sex and other sexual contact with the girl four years ago; she's now 19. Police say the victim later corroborated the incidents, which occurred in a field in South Shenango Township, the Meadville Tribune reports. White was arrested Thursday on multiple charges of unlawful sexual contact with a minor and corruption of minors, and jailed on \$25,000 bond. \"Needless to say, he won't be getting an interview,\" quips the Patriot-News.

    \end{AcademicBox}
    \vspace{-1em}
    \caption{Llama-3.1-70B-Instruct in MultiNews}
    \label{fig:gpt4o_multi_news}
\end{figure*}

\begin{figure*}[t]
    \begin{AcademicBox}[\footnotesize GPT-4o in LoCoMo]

            Passage {[15]} :   
            \begin{verbatim}
...\nCaroline said, \"Painting's great for expressing myself. I 
love creating art! Here's a recent self-portrait I made last we-
ek. \" and shared a photo of a painting of a woman with a blue 
face. \nMelanie said,\"Carolin ...\end{verbatim}

\vspace{-2pt} \hrule \vspace{4pt}
\textbf{\textit{Question:}} \\
When did Caroline draw a self-portrait? \\
\vspace{-5pt} \hrule \vspace{4pt}
\textbf{\textit{Answer:}} \\
Caroline drew a self-portrait last week [15]. \\
\vspace{-5pt} \hrule \vspace{4pt}
\textbf{\textit{Ground truth:}} \\
 The week before 23 August 2023

    \end{AcademicBox}
    \vspace{-1em}
    \caption{GPT-4o in LoCoMo}
    \label{fig:gpt4o_multi_news}
\end{figure*}

\begin{figure*}[t]
    \begin{AcademicBox}[\footnotesize Claude-3.5-Sonnet in LoCoMo]
  Passage {[14]} :   
            \begin{verbatim}
... I'm so stoked about it! Check it out!\" and shared a photo-
graphy of a man with purple hair and glasses taking a selfie.\n
Joanna said, \"Wow, your new hair color looks ...\end{verbatim}

\vspace{-2pt} \hrule \vspace{4pt}
\textbf{\textit{Question:}} \\
What color did Nate choose for his hair? \\
\vspace{-5pt} \hrule \vspace{4pt}
\textbf{\textit{Answer:}} \\
Nate dyed his hair purple, as revealed in a selfie he shared with Joanna [14]. He chose this bright and bold color because he wanted to stand out from regular options, reflecting his personality [14]. \\
\vspace{-5pt} \hrule \vspace{4pt}
\textbf{\textit{Ground truth:}} \\
purple

    \end{AcademicBox}
    \vspace{-1em}
    \caption{Claude-3.5-Sonnet in LoCoMo}
    \label{fig:gpt4o_multi_news}
\end{figure*}

\begin{figure*}[t]
    \begin{AcademicBox}[\footnotesize Phi-3.5-mini-Instruct in LoCoMo]
  Passage {[62]} :   
            \begin{verbatim}
...  dogs love running around the park. But one thing they hate 
is snow. I took them to a snowy one last winter and they were so
confused! They definitely prefer nice, sunny days in the grass.\
"and shared a photo of a dog wearing a shirt standing on a ...\end{verbatim}

\vspace{-2pt} \hrule \vspace{4pt}
\textbf{\textit{Question:}} \\
How do Audrey's dogs react to snow? \\
\vspace{-5pt} \hrule \vspace{4pt}
\textbf{\textit{Answer:}} \\
Audrey's dogs seem to be confused and not particularly fond of snow, as they had a difficult time adjusting during a snowy day visit to the park (Conversation [62]). Reference: Conversation [62] \\
\vspace{-5pt} \hrule \vspace{4pt}
\textbf{\textit{Ground truth:}} \\
 Confused

    \end{AcademicBox}
    \vspace{-1em}
    \caption{Phi-3.5-mini-Instruct in LoCoMo}
    \label{fig:gpt4o_multi_news}
\end{figure*}

\begin{figure*}[t]
    \begin{AcademicBox}[\footnotesize Llama-3.1-8B-Instruct in LoCoMo]
Passage {[15]} :   
            \begin{verbatim}
...   He's even keeping my other pets active.\"\nJoanna said, \"
Pets sure do make life better! Glad Max is bringing you lots of 
joy.\"\n Nate said, \"Yep, totally! Pets make us so much happier 
and never let us down. Have you ...\end{verbatim}

\vspace{-2pt} \hrule \vspace{4pt}
\textbf{\textit{Question:}} \\
 Who was the new addition to Nate's family in May 2022? \\
\vspace{-5pt} \hrule \vspace{4pt}
\textbf{\textit{Answer:}} \\
Nate adopted a dog named Max in May 2022 [15]. \\
\vspace{-5pt} \hrule \vspace{4pt}
\textbf{\textit{Ground truth:}} \\
Max

    \end{AcademicBox}
    \vspace{-1em}
    \caption{Llama-3.1-8B-Instruct in LoCoMo}
    \label{fig:gpt4o_multi_news}
\end{figure*}

\begin{figure*}[t]
    \begin{AcademicBox}[\footnotesize Qwen2-57B-A14B-Instruct in LoCoMo]
Passage {[15]} :   
            \begin{verbatim}
...  Hey Mel, great to chat with you again! So much has happened 
since we last spoke - I went to an LGBTQ conference two days ago 
and it was really special. I got the chance to meet and ...\end{verbatim}

\vspace{-2pt} \hrule \vspace{4pt}
\textbf{\textit{Question:}} \\
What workshop did Caroline attend recently? \\
\vspace{-5pt} \hrule \vspace{4pt}
\textbf{\textit{Answer:}} \\
Caroline attended an LGBTQ conference two days ago [15]. \\
\vspace{-5pt} \hrule \vspace{4pt}
\textbf{\textit{Ground truth:}} \\
LGBTQ+ counseling workshop

    \end{AcademicBox}
    \vspace{-1em}
    \caption{Qwen2-57B-A14B-Instruct in LoCoMo}
    \label{fig:gpt4o_multi_news}
\end{figure*}

\begin{figure*}[t]
    \begin{AcademicBox}[\footnotesize Llama-3.1-70B-Instruct in LoCoMo]
Passage {[13]} :   
            \begin{verbatim}
... Nate said, \"Coconut milk ice cream is at the top of my list.
It's so smooth and creamy with a tropical coconut twist. Plus, 
it's dairy-free for people who can't have lactose or who  ...\end{verbatim}

\vspace{-2pt} \hrule \vspace{4pt}
\textbf{\textit{Question:}} \\
What is Nate's favorite dish from the cooking show he hosted? \\
\vspace{-5pt} \hrule \vspace{4pt}
\textbf{\textit{Answer:}} \\
Nate's favorite dish from the cooking show he hosted is coconut milk ice cream [13]. \\
\vspace{-5pt} \hrule \vspace{4pt}
\textbf{\textit{Ground truth:}} \\
 Coconut milk ice cream

    \end{AcademicBox}
    \vspace{-1em}
    \caption{Llama-3.1-70B-Instruct in LoCoMo}
    \label{fig:gpt4o_multi_news}
\end{figure*}

\begin{figure*}[t]
    \begin{AcademicBox}[\footnotesize GPT-4o in NIAH]
            Passage {[1]} :   
            \begin{verbatim}
... Till recently I thought it didn't, but the truth turns out 
to be  more complicated. Originally, yes, it was meaningless. 
Now it seems to have acquired a meaning.  And yet those who di-
slike the term are probably right, because if it means what I 
think it does, we don't need it The best way to start your day
in New York is to watch the sunrise from the Brooklyn Bridge 
and have a bagel. ...\end{verbatim}

\vspace{-2pt} \hrule \vspace{4pt}
\textbf{\textit{Question:}} \\
What is the best way to start your day in New York? \\
\vspace{-5pt} \hrule \vspace{4pt}
\textbf{\textit{Answer:}} \\
The best way to start your day in New York is to watch the sunrise from the Brooklyn Bridge and then have a bagel [1]. \\
\vspace{-5pt} \hrule \vspace{4pt}
\textbf{\textit{Ground truth:}} \\
watch the sunrise from the Brooklyn Bridge and have a bagel

    \end{AcademicBox}
    \vspace{-1em}
    \caption{GPT-4o in NIAH}
    \label{fig:gpt4o_multi_news}
\end{figure*}

\begin{figure*}[t]
    \begin{AcademicBox}[\footnotesize Claude-3.5-Sonnet in NIAH]
            Passage {[4]} :   
            \begin{verbatim}
... In Paris, the most enjoyable activity is strolling along the 
Seine River and savoring a croissant.  For all practical purpos-
es, ...\end{verbatim}

\vspace{-2pt} \hrule \vspace{4pt}
\textbf{\textit{Question:}} \\
 What is the most enjoyable activity in Paris? \\
\vspace{-5pt} \hrule \vspace{4pt}
\textbf{\textit{Answer:}} \\
According to the passage, the most enjoyable activity in Paris is strolling along the Seine River and savoring a croissant [4].\\
\vspace{-5pt} \hrule \vspace{4pt}
\textbf{\textit{Ground truth:}} \\
strolling along the Seine River and savoring a croissant

    \end{AcademicBox}
    \vspace{-1em}
    \caption{Claude-3.5-Sonnet in NIAH}
    \label{fig:gpt4o_multi_news}
\end{figure*}

\begin{figure*}[t]
    \begin{AcademicBox}[\footnotesize Phi-3.5-mini-Instruct in NIAH]
Passage {[1]} :   
            \begin{verbatim}
... In Bangkok, the top thing to do is to visit the Grand Palace
and sample street food.. A top-ranked professional chess ...\end{verbatim}

\vspace{-2pt} \hrule \vspace{4pt}
\textbf{\textit{Question:}} \\
What is the top thing to do in Bangkok?  \\
\vspace{-5pt} \hrule \vspace{4pt}
\textbf{\textit{Answer:}} \\
Visit the Grand Palace and sample street food [1].\\
\vspace{-5pt} \hrule \vspace{4pt}
\textbf{\textit{Ground truth:}} \\
visit the Grand Palace and sample street food

    \end{AcademicBox}
    \vspace{-1em}
    \caption{Phi-3.5-mini-Instruct in NIAH}
    \label{fig:gpt4o_multi_news}
\end{figure*}

\begin{figure*}[t]
    \begin{AcademicBox}[\footnotesize Llama-3.1-8B-Instruct in NIAH]
            Passage {[1]} :   
            \begin{verbatim}
... At the extreme end of the spectrum are crack and meth The 
best thing to do in Beijing is to take a walk in Chaoyang Park
and have a cup of Espresso in the evenin ...\end{verbatim}

\vspace{-2pt} \hrule \vspace{4pt}
\textbf{\textit{Question:}} \\
What is the best thing to do in Beijing ?  \\
\vspace{-5pt} \hrule \vspace{4pt}
\textbf{\textit{Answer:}} \\
The best thing to do in Beijing is to take a walk in Chaoyang Park and have a cup of Espresso in the evening [1].\\
\vspace{-5pt} \hrule \vspace{4pt}
\textbf{\textit{Ground truth:}} \\
take a walk in Chaoyang Park and have a cup of Espresso in the evening

    \end{AcademicBox}
    \vspace{-1em}
    \caption{Llama-3.1-8B-Instruct in NIAH}
    \label{fig:gpt4o_multi_news}
\end{figure*}

\begin{figure*}[t]
    \begin{AcademicBox}[\footnotesize Qwen2-57B-A14B-Instruct in NIAH]
            Passage {[10]} :   
            \begin{verbatim}
... the must-do is to explore Shinjuku Gyoen National Garden and 
taste matcha tea\end{verbatim}

\vspace{-2pt} \hrule \vspace{4pt}
\textbf{\textit{Question:}} \\
 What is the must-do in Tokyo?   \\
\vspace{-5pt} \hrule \vspace{4pt}
\textbf{\textit{Answer:}} \\
The must-do in Tokyo is to explore Shinjuku Gyoen National Garden and taste matcha tea [10]. \\
\vspace{-5pt} \hrule \vspace{4pt}
\textbf{\textit{Ground truth:}} \\
explore Shinjuku Gyoen National Garden and taste matcha tea

    \end{AcademicBox}
    \vspace{-1em}
    \caption{Qwen2-57B-A14B-Instruct in NIAH}
    \label{fig:gpt4o_multi_news}
\end{figure*}

\begin{figure*}[t]
    \begin{AcademicBox}[\footnotesize Llama-3.1-70B-Instruct in NIAH]
            Passage {[26]} :   
            \begin{verbatim}
... thanks to the internet, there The most effective way to lose 
weight is by eating a balanced diet and exercising regularly.'s 
a path. Anyon...\end{verbatim}

\vspace{-2pt} \hrule \vspace{4pt}
\textbf{\textit{Question:}} \\
What is the most effective way to lose weight?   \\
\vspace{-5pt} \hrule \vspace{4pt}
\textbf{\textit{Answer:}} \\
The most effective way to lose weight is by eating a balanced diet and exercising regularly [26]. \\
\vspace{-5pt} \hrule \vspace{4pt}
\textbf{\textit{Ground truth:}} \\
 by eating a balanced diet and exercising regularly.

    \end{AcademicBox}
    \vspace{-1em}
    \caption{Llama-3.1-70B-Instruct in NIAH}
    \label{fig:gpt4o_multi_news}
\end{figure*}

\begin{figure*}[t]
    \begin{AcademicBox}[\footnotesize GPT-4o in Counting Stars]
            Passage {[97]} :   
            \begin{verbatim}
The little penguin counted 36 \u2605\n not speed up.  Essays sh-
ould do the opposite.  Essays should aim for maximum surprise.I 
was afraid of flying for a long time and could only travel vica-
riously...\end{verbatim}

            Passage {[193]} :   
            \begin{verbatim}
Socially they \nThe little penguin counted 76 \u2605\n'd seem 
more like rants posted by nutjobs on their GeoCities sites than 
the genteel and beautifully typeset compositions published in 
The New Yorker...
\end{verbatim}

\vspace{-2pt} \hrule \vspace{4pt}
\textbf{\textit{Question:}} \\
Please help the little penguin collect the number of \(\star\).   \\
\vspace{-5pt} \hrule \vspace{4pt}
\textbf{\textit{Answer:}} \\
\{"little\_penguin": [36, 76], "passage\_id": [97, 193]\} \\
\vspace{-5pt} \hrule \vspace{4pt}
36, 76

    \end{AcademicBox}
    \vspace{-1em}
    \caption{GPT-4o in Counting Stars}
    \label{fig:gpt4o_multi_news}
\end{figure*}

\begin{figure*}[t]
    \begin{AcademicBox}[\footnotesize Claude-3.5-Sonnet in Counting Stars]
            Passage {[18]} :   
            \begin{verbatim}
Or that \nThe little penguin counted 58 \u2605\n's what it feels 
like. What's really going on is that you have different expectat
...
\end{verbatim}

            Passage {[36]} :   
            \begin{verbatim}
but don't put bars on the windows. \nThe little penguin counted 
106 \u2605\n[2] Even now I sometimes write essays that are not 
meant ...
\end{verbatim}

\vspace{-2pt} \hrule \vspace{4pt}
\textbf{\textit{Question:}} \\
Please help the little penguin collect the number of \(\star\).   \\
\vspace{-5pt} \hrule \vspace{4pt}
\textbf{\textit{Answer:}} \\
\{"little\_penguin": [58, 106], "passage\_id": [18, 36]\} \\
\vspace{-5pt} \hrule \vspace{4pt}
 58, 106

    \end{AcademicBox}
    \vspace{-1em}
    \caption{Claude-3.5-Sonnet in Counting Stars}
    \label{fig:gpt4o_multi_news}
\end{figure*}

\begin{figure*}[t]
    \begin{AcademicBox}[\footnotesize Phi-3.5-mini-Instruct in Counting Stars]
            Passage {[18]} :   
            \begin{verbatim}
... which is the reason they   began blogging in the first place 
\nThe little penguin counted 16 \u2605\n....
\end{verbatim}

            Passage {[114]} :   
            \begin{verbatim}
... startup hubs is the degree to which people help one another 
out \nThe little penguin counted 34 ...
\end{verbatim}

\vspace{-2pt} \hrule \vspace{4pt}
\textbf{\textit{Question:}} \\
Please help the little penguin collect the number of \(\star\).   \\
\vspace{-5pt} \hrule \vspace{4pt}
\textbf{\textit{Answer:}} \\
\{"little\_penguin": [58, 106], "passage\_id": [18, 36]\} \\
\vspace{-5pt} \hrule \vspace{4pt}
16, 106

    \end{AcademicBox}
    \vspace{-1em}
    \caption{Phi-3.5-mini-Instruct in Counting Stars}
    \label{fig:gpt4o_multi_news}
\end{figure*}

\begin{figure*}[t]
    \begin{AcademicBox}[\footnotesize Llama-3.1-8B-Instruct in Counting Stars]
            Passage {[1]} :   
            \begin{verbatim}
July 2010What hard liquor \nThe little penguin counted 65 \u2605
\n, cigarettes, heroin, and crack have....
\end{verbatim}

            Passage {[58]} :   
            \begin{verbatim}
... We would end up getting all the users \nThe little penguin 
counted 58  ...
\end{verbatim}

\vspace{-2pt} \hrule \vspace{4pt}
\textbf{\textit{Question:}} \\
Please help the little penguin collect the number of \(\star\).   \\
\vspace{-5pt} \hrule \vspace{4pt}
\textbf{\textit{Answer:}} \\
\{"little\_penguin": [65, 58], "passage\_id": [1, 58]\} \\
\vspace{-5pt} \hrule \vspace{4pt}
65, 58

    \end{AcademicBox}
    \vspace{-1em}
    \caption{Llama-3.1-8B-Instruct in Counting Stars}
    \label{fig:gpt4o_multi_news}
\end{figure*}

\begin{figure*}[t]
    \begin{AcademicBox}[\footnotesize Qwen2-57B-A14B-Instruct in Counting Stars]
  Passage {[1]} :   
            \begin{verbatim}
July 2010What hard liquor \nThe little penguin counted 65 \u2605
\n, cigarettes, heroin, and crack have....
\end{verbatim}

            Passage {[58]} :   
            \begin{verbatim}
... We would end up getting all the users \nThe little penguin
counted 58  ...
\end{verbatim}

\vspace{-2pt} \hrule \vspace{4pt}
\textbf{\textit{Question:}} \\
Please help the little penguin collect the number of \(\star\).   \\
\vspace{-5pt} \hrule \vspace{4pt}
\textbf{\textit{Answer:}} \\
\{"little\_penguin": [65, 58], "passage\_id": [1, 58]\} \\
\vspace{-5pt} \hrule \vspace{4pt}
65, 58

    \end{AcademicBox}
    \vspace{-1em}
    \caption{Qwen2-57B-A14B-Instruct in Counting Stars}
    \label{fig:gpt4o_multi_news}
\end{figure*}

\begin{figure*}[t]
    \begin{AcademicBox}[\footnotesize Llama-3.1-70B-Instruct in Counting Stars]
            Passage {[18]} :   
            \begin{verbatim}
... Or that \nThe little penguin counted 58 \u2605\n's what it 
feels like ....
\end{verbatim}

            Passage {[36]} :   
            \begin{verbatim}
but don't put bars on the windows. \nThe little penguin counted 
106 \u2605\n[2] Even now I sometimes write  ...
\end{verbatim}

\vspace{-2pt} \hrule \vspace{4pt}
\textbf{\textit{Question:}} \\
Please help the little penguin collect the number of \(\star\).   \\
\vspace{-5pt} \hrule \vspace{4pt}
\textbf{\textit{Answer:}} \\
\{"little\_penguin": [58, 106], "passage\_id": [18, 36]\} \\
\vspace{-5pt} \hrule \vspace{4pt}
58, 106
\end{AcademicBox}
\vspace{-1em}
\caption{Llama-3.1-70B-Instruct in Counting Stars}
\label{fig:gpt4o_multi_news}
\end{figure*}

\end{document}